%% file: main.tex
\documentclass[11pt]{article}
\usepackage{deauthor,times,graphicx}

\input{commands}

\makeatletter           

\begin{document}

\title{Rethinking Benchmarks for Differentially Private Image Classification\thanks{Authors SMok, SK, and SMoh have equal contributions and are listed alphabetically in order of \emph{first} name. Authors FT and GK are listed in reverse alphabetical order.}}

\author{
  Sabrina Mokhtari\\
  University of Waterloo\\
  \texttt{s4mokhtari@uwaterloo.ca} \\
  \and
  Sara Kodeiri \\
  University of Waterloo \\
  \texttt{skodeiri@uwaterloo.ca} \\
\and
  Shubhankar Mohapatra \\
  University of Waterloo \\
  \texttt{s3mohapa@uwaterloo.ca} \\
  \and
  Florian Tram\`er \\
  ETH Z\"urich \\
  \texttt{florian.tramer@inf.ethz.ch} \\
  \and
  Gautam Kamath \\
  University of Waterloo \\
  \texttt{g@csail.mit.edu} \\
}

\maketitle

\begin{abstract}
We revisit benchmarks for differentially private image classification. We suggest a comprehensive set of benchmarks, allowing researchers to evaluate techniques for differentially private machine learning in a variety of settings, including with and without additional data, in convex settings, and on a variety of qualitatively different datasets. We further test established techniques on these benchmarks in order to see which ideas remain effective in different settings. Finally, we create a publicly available leader board for the community to track progress in differentially private machine learning.
\end{abstract}

\input{sections/1_intro}
\input{sections/2_benchdes}
\input{sections/3_datamethods}
\input{sections/4_exp}

\input{sections/5_results}
\input{sections/6_relatedfuture}

\input{sections/7_conclusion}

\bibliographystyle{abbrv}
\bibliography{biblio}  

\appendix

\end{document}

%% file: commands.tex
\newcommand{\eat}[1]{}

\usepackage{latexsym}
\usepackage{amsfonts}
\usepackage{amsmath}
\usepackage{amssymb}
\usepackage{xcolor}
\usepackage{colortbl}
\usepackage{epsfig}
\usepackage{xspace}
\usepackage{graphicx}
\usepackage{subcaption}
\usepackage{tablefootnote}
\usepackage{paralist}
\usepackage{enumerate}
\usepackage{hyperref}
\usepackage{xy}
\usepackage{comment}
\usepackage{booktabs}
\usepackage{balance}
\usepackage{stmaryrd}
\usepackage{pifont}
\usepackage{hhline}
\usepackage{listings}
\usepackage{array}
\usepackage{float}
\usepackage{threeparttable}

\usepackage{amsthm}
\usepackage{wrapfig}

\newcolumntype{L}[1]{>{\raggedright\let\newline\\\arraybackslash\hspace{0pt}}m{#1}}
\newcolumntype{C}[1]{>{\centering\let\newline\\\arraybackslash\hspace{0pt}}m{#1}}
\newcolumntype{R}[1]{>{\raggedleft\let\newline\\\arraybackslash\hspace{0pt}}m{#1}}

\usepackage{epsfig}
\usepackage{multirow}
\usepackage{url}

\usepackage{tikz}
\usetikzlibrary{calc}

\usepackage{algorithm2e}

\SetCommentSty{mycommfont}
\usepackage{adjustbox}

\newtheorem{myDefinition}{\textbf{Definition}}

\newcommand{\bi}{\begin{itemize}}
	\newcommand{\ei}{\end{itemize}}

\newcommand{\be}{\begin{enumerate}}
	\newcommand{\ee}{\end{enumerate}}
\newcommand{\beqn}{\begin{eqnarray*}}
	\newcommand{\eeqn}{\end{eqnarray*}}

\renewcommand{\epsilon}{\varepsilon}

\newcounter{ccc}

\newcommand{\eop}{\hspace*{\fill}\mbox{$\Box$}\vspace{1ex}}     

\newcommand{\nthesection}{\arabic{section}}

\newcounter{alg}[section]
\renewcommand{\thealg}{\nthesection.\arabic{alg}}

\newcounter{arule}
\renewcommand{\thearule}{\arabic{arule}}

\makeatletter
\newcommand\figcaption{\def\@captype{figure}\caption}
\newcommand\tabcaption{\def\@captype{table}\caption}
\makeatother

\definecolor{shadecolor}{RGB}{200,200,200}

\definecolor{shadecolor1}{RGB}{230,230,230}

\definecolor{shadecolor1}{RGB}{255, 114, 118}

\tikzstyle{mybox} = [draw=black, fill=black!5, thick,
rectangle, rounded corners, inner sep=0pt, inner ysep=6pt]
\tikzstyle{fancytitle} =[fill=black, text=white]

\NewDocumentCommand{\yang}{ mO{} }{\textcolor{blue}{\textsuperscript{\textit{yang}}\textsf{\small[#1]}}}

%% file: sections/1_intro.tex
\section{Introduction}\label{sec:intro}

Machine learning (ML) models have been repeatedly demonstrated to leak sensitive information pertaining to their training data. 
These issues manifest through a number of different types of attacks, including membership inference~\cite{HomerSRDTMPSNC08,ShokriSSS17}, model inversion~\cite{FrederiksonJR15}, and even training data extraction~\cite{CarliniTWJHLRBSEOR21,CarliniHNJSTBIW23, SomepalliSGGG23}. 
This can be problematic if the training data contains privacy-sensitive information belonging to people.
To alleviate such concerns, a popular solution is differential privacy (DP)~\cite{DworkMNS06}.
DP is a rigorous notion of individual data privacy, which can be used to mask the presence or absence of any single training data point when observing a trained model.
In particular, training a model with DP provably prevents all the aforementioned attacks. 

The past decade has seen significant effort and success in training ML models with DP, including image classifiers~\cite{AbadiCGMMTZ16,TramerB21,PapernotTSCE21,DeBHSB22}, large language models~\cite{AnilGGKM21,YuNBGIKKLMWYZ22,LiTLH22}, and other generative models~\cite{XieLWWZ18,BeaulieuJonesWWLBBG19,CaoBVFK21,BieKZ23,DockhornCVK23,HarderJSP23}. 
However, in a recent position paper, Tram\`er, Kamath, and Carlini critique a number of trends in DP ML~\cite{TramerKC24}.
Most pertinent to our work, they question whether benchmarks used in DP ML are truly measuring progress in the field, specifically in the context of DP image classification, which will be our focus. 
The most common benchmark datasets used in DP image classification include MNIST~\cite{LeCunCB10}, CIFAR-10~\cite{Krizhevsky09}, and ImageNet~\cite{DengDSLLF09}.
While significant progress has been made on each, TKC question whether this progress generalizes to privacy-sensitive settings where DP may be deployed.
For example, CIFAR-10 and ImageNet are both composed primarily of natural images of everyday objects.
While these datasets indeed have some privacy concerns~\cite{BirhaneP21}, it is less clear whether they resemble domains where DP is of high practical concern, such as, e.g., medical images.
Since, informally speaking, medical images appear to qualitatively differ from those in the aforementioned datasets, it is unclear whether techniques previously established to be effective remain so in these settings. 
This question is even more pronounced when models are pre-trained on public data (i.e., supplementary data which is not subject to any privacy constraints), a popular trend in private ML.
In such settings, the chosen ``public'' datasets are often visually similar to the private ones -- as a representative example,~\cite{DeBHSB22} treat ImageNet as public and privately fine-tune on CIFAR-10. 
On the other hand, for domains such as medical images, private images may be specialized and ill-represented in public pre-training datasets.
Finally, further muddying the waters is the fact that results on these benchmark datasets are often reported for incomparable settings, in particular, with vastly differing public pre-training datasets.
Overall, these issues make it difficult to isolate which ideas and techniques are truly effective in privacy-critical settings.

Our contributions are as follows:
\begin{itemize}
\item We propose standardized benchmark datasets and evaluation settings to measure progress in DP image classification, with a particular focus on privacy-sensitive domains;
\item We release a public leaderboard for DP ML, for the community to track improvements on these benchmarks;
\item We evaluate previously established techniques for DP image classification across a variety of settings to see which are and are not broadly effective. 
\end{itemize} 

\section{Preliminaries}
We recall the celebrated notion of differential privacy. 
\begin{myDefinition}[\cite{DworkMNS06,DworkKMMN06}]
 An algorithm $M\ : \mathcal{X}^n \rightarrow \mathcal{Y}$ is $(\varepsilon, \delta)$-differentially private if, for all neighboring datasets (i.e., datasets that differ in exactly one entry) $X$ and $X'$ and all events $S \subseteq \mathcal{Y}$, we have that 
    $\Pr[M(X) \in S] \leq e^\varepsilon \Pr[M(X') \in S] + \delta$.
\end{myDefinition}
DP is a quantitative definition of individual data privacy. The privacy cost is measured by the parameters ($\epsilon,\delta$), also called the privacy budget. Smaller values of $\epsilon$ correspond to stricter privacy guarantees, and it is standard in the literature to set $\delta \ll \frac{1}{n}$, where $n$ is the size of the database.
Complex DP algorithms can be built from the basic algorithms following two important properties of differential privacy:
1) Post-processing states that for any function $g$ defined over the output of the mechanism $\mathcal{M}$, if $\mathcal{M}$ satisfies ($\epsilon,\delta$)-DP, so does $g(\mathcal{M})$;
2) Basic composition states that if for each $i \in [k]$, mechanism $\mathcal{M}_i$ satisfies ($\epsilon_i,\delta_i$)-DP, 
then a mechanism sequentially applying $\mathcal{M}_1$, $\mathcal{M}_2, \ldots, \mathcal{M}_k$ satisfies ($\sum_{i=1}^k\epsilon_i, \sum_{i=1}^k\delta_i$)-DP.

Given a function $f:\mathcal{D} \rightarrow \mathbb{R}^d$, the \emph{Gaussian mechanism} adds noise drawn from a normal distribution $\mathcal{N}(0,S_f^2\sigma^2)$  to each dimension of the output, where $S_f$
is the $\ell_2$-sensitivity of $f$, 
defined as
$S_f = \max_{D,D'\text{differ in a row}} \|f(D) - f(D')\|_2$.
For $\epsilon\in (0,1)$, if $\sigma \geq \sqrt{2\ln(1.25/\delta)}/\epsilon$, 
then the Gaussian mechanism satisfies $(\epsilon,\delta)$-DP.

We focus on training ML models subject to DP, which (due to its post-processing property) allows the trained model to be publicly released without further privacy concerns.
The most popular method for DP training of ML models is differentially private stochastic gradient descent (DPSGD)~\cite{SongCS13,BassilyST14,AbadiCGMMTZ16}.
In contrast to non-private SGD where batches are sliced from the training dataset, DPSGD at each iteration works by sampling ``lots'' from the training with probability $L/n$, where $L$ is the (expected) lot size and $n$ is the total data size. A set of queries are computed over those samples. These queries include gradient computation, updates to batch normalization, or accuracy metric calculations. As there is no a priori bound on these query outputs, the sensitivity $S_f$ is set by clipping the maximum $\ell_2$ norm of the gradient to a user-defined parameter $C$. The gradient of each point is then noised and published. All DP optimizers follow the same framework in which they take steps on the computed noisy gradient as in its non-private counterpart. The privacy cost of the whole training procedure is calculated using privacy accounting techniques. We discuss the specifics of DPSGD for our experiments in Section~\ref{sec:BenchDes}.

%% file: sections/2_benchdes.tex
\section{Benchmark design}\label{sec:BenchDes}

In this section, we report our specific prescriptions for benchmarks, including datasets, parameters, and best practices, in a variety of settings, in order to standardize and (ideally) propel progress in DP image classification in privacy-critical settings.
We note that we (intentionally) do not introduce any new datasets, and instead appeal to existing ones.
This is because using established datasets allows for easier comparisons between the private and non-private setting, and introducing an entirely new dataset would serve no benefit for our setting.

\paragraph{Datasets} \label{data}
We prescribe using the following two medical image datasets (which have been commonly used in other areas of machine learning) as benchmarks for DP ML: a) CheXpert~\cite{IrvinRKYCCMHBSSMHSJLLPLN19}, a chest X-ray dataset; and b) EyePACS~\cite{eyepacs}, a diabetic retinopathy dataset.
These datasets are primarily chosen due their privacy-critical domain.
We hope that progress on these benchmarks would align with progress (i.e., increased utility) on truly private tasks in such settings.
Secondarily, we choose these datasets due to diversity in their sizes, balance of classes, and in the case of CheXpert, for inclusion of a multilabel dataset. 
Further description of these datasets and justification of these choices appears in Section~\ref{sec:datasets_methods}. 
In addition, we recommend continuing to use CIFAR-10~\cite{Krizhevsky09} and ImageNet~\cite{DengDSLLF09} as benchmarks for training DP ML models \emph{from scratch}, without any pretraining data. 
Indeed, keeping the caveats of~\cite{TramerKC24} in mind, the popularity of these datasets still allows for direct comparison of accuracy on these tasks, and thus to track ``how far behind'' DP ML is behind the non-private setting. 

\paragraph{Public datasets} One of the most successful ways to improve the utility of DP ML has been pre-training the model on ``public'' data (i.e., data free of any privacy constraints). 
As discussed by~\cite{TramerKC24}, the size and nature of the pre-training data can dramatically affect the downstream utility of a privately fine-tuned model. 
Therefore, for fair comparison between different techniques, we prescribe tracking progress with the following datasets treated as public: a) no public data, for the ``purest'' measure of progress in DP ML; b) ImageNet-1K, perhaps the most commonly used large image classification dataset c) LAION-2B, due to it being the pre-training data for OpenCLIP's ViT-G/14 (representing the common use-case of privately fine-tuning a pre-trained CLIP model), and d) ``anything goes.'' 
To elaborate on the last of these, we use ``anything goes'' to refer to the case when public pre-training data is unrestricted (barring data-leakage-like considerations where the private dataset contaminates the public one): it may include large-scale Internet datasets, additional domain-specific data, etc. 
As mentioned before, results in this category may not be directly comparable with each other.
Nonetheless, they serve as a measure of absolute progress on a benchmark. 

\paragraph{Privacy parameters} It is not clear how to compare results on DP image classification at varying levels of the privacy parameters $\varepsilon$ and $\delta$. 
For example, is $90\%$ accuracy at $\varepsilon = 1$ better or worse than $95\%$ at $\varepsilon = 2$?
We propose fixing the value of $\varepsilon$ to be $1, 3, 5$ and $8$ to facilitate direct comparisons between results.
This set of $\varepsilon$ covers both high and low privacy regimes across the range usually considered in DP ML.
We additionally propose fixing $\delta$ to to be the largest power of 10 that is at most the inverse of the training set size (consistent with previous parameter settings), though in many parameter regimes, $\delta$ can be dramatically increased or decreased with minor effect on the value of $\varepsilon$.

\paragraph{Privacy accounting.} Every DP algorithm is associated with a proof of privacy, which provides an upper bound on the value of $\varepsilon$ and $\delta$.
For DPSGD, this is generally automated using ``privacy accountants,'' which take as input various hyperparameters and $\delta$, and outputs the value of $\varepsilon$.
Over time, improved accounting methods have given increasingly tight analyses, culminating in ``exact'' privacy accounting techniques~\cite{AbadiCGMMTZ16,Mironov17,MironovTZ19,KoskelaJH20,GopiLW21}.
However, as highlighted by some recent works~\cite{PonomarevaHKXDMVCT23,ChuaGKKMSZ24,LebedaRKS24}, simply using a tighter accountant may give the illusion of an improved result, even if the training procedure is identical.
Therefore, we recommend that the privacy accounting method (or, if not using DPSGD, the specific proof followed) is reported in order to keep track of such discrepancies (ideally, all future DPSGD works ought to use exact privacy accountants).

\paragraph{Applicable techniques.} The most popular algorithm for DP ML is DPSGD~\cite{SongCS13,BassilyST14,AbadiCGMMTZ16}, in part due to its flexibility: it can be used to privately train any differentiable model, even non-convex ones.
Other methods, such as objective perturbation~\cite{ChaudhuriMS11,KiferST12,IyengarNSTTW19, RedbergKW23}, are usually applicable only to convex models. 
Consequently, in addition to several non-convex settings, we suggest some standardized convex settings so that a wider variety of methods may be compared and evaluated. 
We recommend linear probe (i.e., logistic regression) on a) Wide ResNet-28-10 pre-trained on ImageNet-1K;\footnote{Inspired by~\cite{DeBHSB22}. While they release their weights in JAX, we release comparable PyTorch weights with the code \href{https://github.com/mshubhankar/DP-Benchmarks}{https://github.com/mshubhankar/DP-Benchmarks}.} and b) OpenCLIP's ViT-G/14 pre-trained on LAION-2B.

\paragraph{``Anything goes'' zero-shot} Parallel to the literature on DP ML, the general ML community has studied the challenging ``zero-shot'' setting, in which goal is to correctly classify a test image without seeing a single image in its training set.
Naturally, this requires large-scale public pre-training to achieve acceptable results.
In terms of DP, this corresponds to $\varepsilon = 0$ but with ``anything goes'' pre-training (described above). 
We suggest tracking the current SOTA for such settings, as a) they serve as an important measure of absolute progress on benchmarks; and b) it is otherwise easy to report a DP result with ``anything goes'' public data and $\varepsilon > 0$ as SOTA, despite being already dominated by existing zero-shot results. 

Overall, we remind that our community's goal ought \emph{not} be to get the highest numbers on these specific datasets, but instead to improve our techniques and understanding of DP image classification for settings that may generalize to those used in practice. 
We thus focus on a breadth of settings to hopefully cover a range of conditions in which DP classifiers may be deployed.
Even if a model can achieve high utility on a benchmark in the ``anything goes'' zero-shot setting, this does not mean the problem is necessarily ``solved.'' 
For instance, due to legal, ethical, computational, or safety reasons, depending on the specific setting, it may not be possible to use large, uncurated public datasets for pre-training in a real-world deployment.
Therefore, we consider all settings outlined above to be of potential practical or technical interest, and do not identify any of them as ``canonical'' or more important than another.

\subsection{Leaderboard}\label{lead}
Tracking progress on benchmark datasets via leaderboards is an established practice in (non-private) ML.\footnote{See, e.g., \url{https://paperswithcode.com/sota}}
This is not yet the case for DP ML: a broad and up-to-date knowledge of the literature is required to keep track of the latest results, making entering the field especially challenging and intimidating for newcomers. 
As one of our contributions, we alleviate this issue by creating and maintaining a leaderboard for DP ML.\footnote{Our leaderboard is available at \href{https://private-machinelearning.github.io/}{https://private-machinelearning.github.io/}}

Due to the particulars of the DP setting, it is unnatural to simply incorporate results into an existing leaderboard for the non-private setting.
Specifically, beyond just the specific dataset, a leaderboard for DP ML would need to track many of the considerations already discussed, including the privacy parameters ($\varepsilon, \delta$), which privacy accountant was used, and which public datasets were used.
Another difference from the non-private setting is the issue of \emph{correctness}.
For a proposed algorithm, the DP guarantees must be mathematically \emph{proven}, and a claimed result could be false if there is a bug in the proof. 
This is in addition to existing concerns from the non-private setting on whether results are independently reproducible or not.
However, since it is notoriously easy to have bugs in a proof of DP, we incorporate a \emph{verification} system to our leaderboard.
By default, all results are unverified when added.
However, anyone is able to submit a pull request to our GitHub to verify that they reproduced the result, and believe correctness of the privacy proof (if applicable).

At present, our leaderboard focuses exclusively on DP image classification (as does this paper), though it may be extended to other problems (e.g., DP natural language understanding or generation). 

%% file: sections/3_datamethods.tex
\begin{figure*}[ht]
\centering
\begin{subfigure}{.24\linewidth}
    \includegraphics[width=\linewidth]{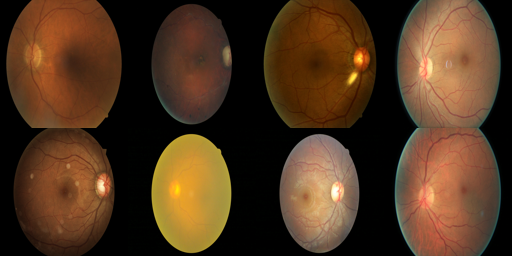}
  \caption{EyePACS}
\end{subfigure}
\begin{subfigure}{.24\linewidth}
    \includegraphics[width=\linewidth]{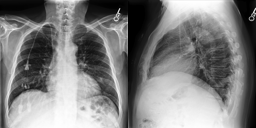}
  \caption{CheXpert}
\end{subfigure}
\begin{subfigure}{.24\linewidth}
    \includegraphics[width=\linewidth]{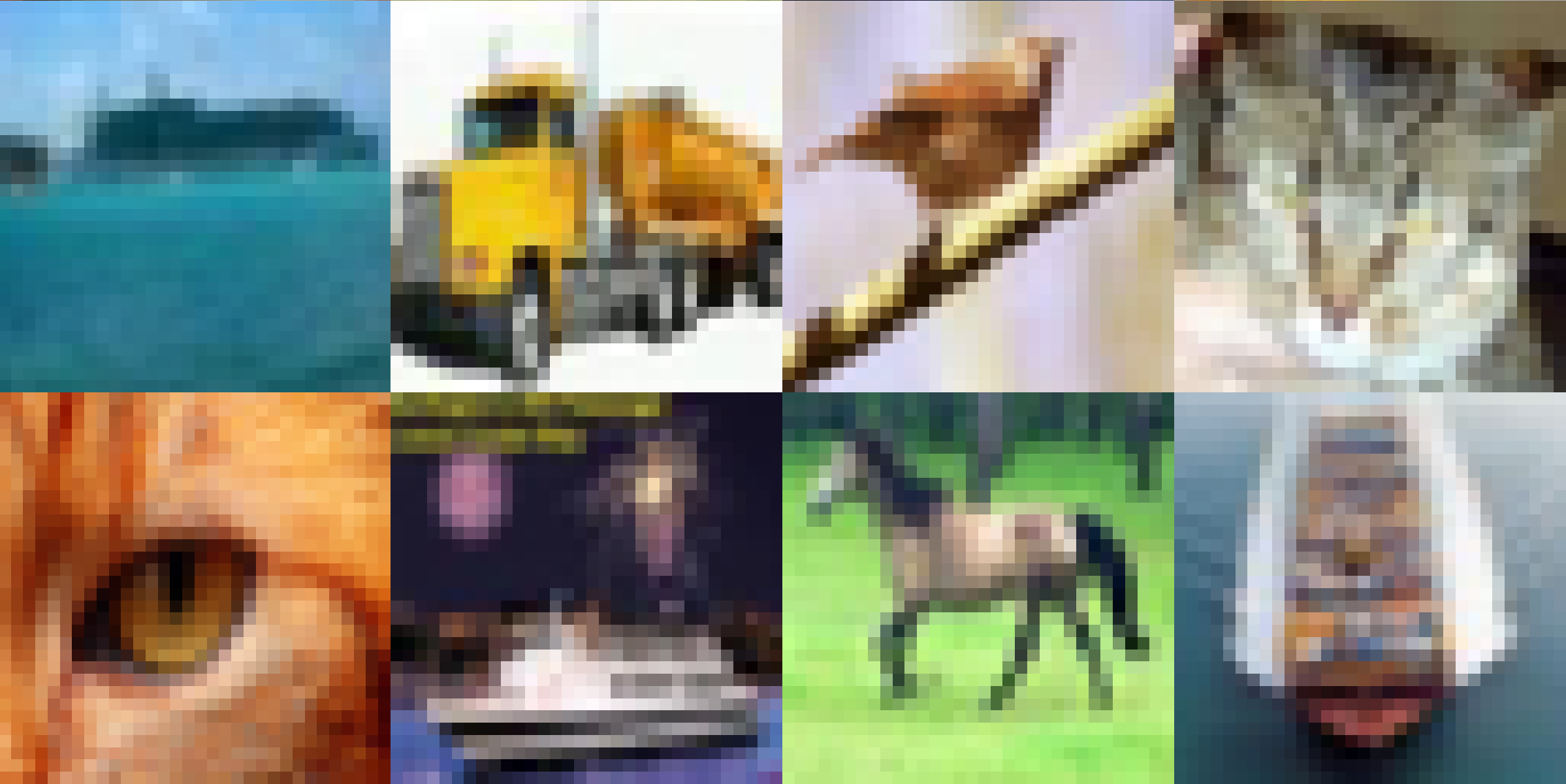}
  \caption{CIFAR-10}
\end{subfigure}
  \begin{subfigure}{.24\linewidth}
    \includegraphics[width=\linewidth]{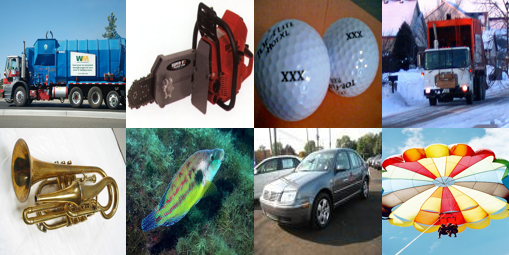}
  \caption{ImageNet}
\end{subfigure}
  \caption{EyePACS and CheXpert qualitatively look different than common benchmark datasets such as CIFAR-10 and ImageNet.}
  \label{fig:datasets}
\end{figure*}

\section{Datasets and Architectures}\label{sec:datasets_methods}
Here, we describe the relevant datasets and architectures, which are later explored in the experiment section.\footnote{Any omitted hyperparameter or architectural details appear in the code repository \href{https://github.com/mshubhankar/DP-Benchmarks}{https://github.com/mshubhankar/DP-Benchmarks}}

\subsection{Overview of datasets}

\paragraph{CheXpert} The CheXpert dataset~\cite{IrvinRKYCCMHBSSMHSJLLPLN19} has 224,316 chest X-ray images of size 390$\times$320 from 64,540 patients. 
Images may have multiple labels, where the possible labels correspond to five pathology classes: `Cardiomegaly', `Edema', `Consolidation', `Atelectasis', and `Pleural Effusion'. 
In our work, following prior state-of-the-art training, we re-scale all images to size 224$\times$224 and augment the dataset using random affine transformations~\cite{yuan2021large, yuan2023libauc}.

    
\paragraph{EyePACS} Kaggle EyePACS ~\cite{eyepacs}
contains retinal images of diverse populations with various degrees of diabetic retinopathy (DR). Each image is classified into one of five classes depending on the severity of the disease. 
The classification task is diagnostic of DR, as measured on a scale from 0 (no DR) to 4 (severe DR). The training set consists of 35,126 and the test set contains 53,576 color eye fundus images.

To speak to these particular dataset selections: as mentioned before, we chose medical images to address a privacy-critical setting where DP may be deployed.
Within this area, chest X-rays and fundus images are two of the most common domains, so we chose one of the most popular datasets from each of these domains.
Additionally, we took guidance from~\cite{raghu2019transfusion}, which also focuses on medical image classification, and studies CheXpert and a Google-proprietary DR dataset.
While there are several public fundus photography datasets, most of them are very small ($< 100$ images) and thus not settings we would expect DP to function well: EyePACS is the most popular one of an acceptable size.

\subsection{Overview of architectures and techniques}

\paragraph{ScatterNets} ScatterNets \cite{Oyallon14} (SN) are convolutional neural networks (CNNs) that utilize pre-defined wavelets for their architecture and filters. 
In other words, the features are ``hand-crafted'' rather than learned from data, and thus use neither public nor private data.
Tram\`er and Boneh \cite{TramerB21} employ this architecture for DP image classification, using DPSGD to train either linear or convolutional layers acting on these features, and demonstrate compelling results on MNIST and CIFAR-10, particularly for small values of $\varepsilon$. 
We exclusively use ScatterNets without any public data. 


\paragraph{Wide-ResNets}
The Wide-ResNet \cite{DBLP:journals/corr/ZagoruykoK16} (WRN) is a variant of the ResNet\cite{he2015deep} that reduces issues of vanishing and exploding gradients by making the model wider instead of deeper. 
De et al.~\cite{DeBHSB22} use them to reach DP SOTA in multiple settings on CIFAR-10.
They consider both DP training from scratch, and DP fine-tuning after being (publicly) pre-trained on ImageNet-1K (downsampled to $32 \times 32$, which we call IN-32~\cite{DBLP:journals/corr/ChrabaszczLH17}).\footnote{They use WRN-16-4 and WRN-40-4 for from-scratch experiments and WRN-28-10 for fine-tuning experiments. For simplicity, we use WRN-28-10 in all our experiments.}
To allow direct comparison, we emulate their setting as much as possible, e.g., using weight standardization~\cite{brock2021}, group normalization, and their choices of hyperparameters for pre-training.
We use both without any public data, and pre-trained on ImageNet-1K.

Additionally, Tang et al.~\cite{tang2023differentiallyprivateimageclassification} utilize WRN-16-4 to achieve DP SOTA performance on CIFAR-10, when no extra public data is used for pretraining. They leverage image priors generated by random processes~\cite{BaradadJurjoWWIT21} instead of starting from random initialization, outperforming \cite{TramerB21} and \cite{DeBHSB22} when they only train from scratch. Moreover, they achieve SOTA performance using only a linear probe, making for a direct comparison to the linear ScatterNet method of \cite{TramerB21}. We adopt the same architecture and replicate their settings to the greatest extent possible, incorporating techniques such as augmentation multiplicity and normalization. Tang et al.~\cite{tang2023differentiallyprivateimageclassification} build on the approach of De et al.~\cite{DeBHSB22} by using the third-to-last layer of the network, which has a dimension of 4096. We adopt a similar strategy but reduce the dimensionality to 2048. This adjustment is necessary due to the larger image sizes in our datasets (CheXpert and EyePACS with 224 $\times$ 224 images) compared to CIFAR-10 (32 $\times$ 32 images) and resource constraints.


\paragraph{CLIP-based models}
CLIP~\cite{radford2021learning} is a popular contrastive learning pre-training technique, which allows one to jointly train a language and image encoder.
CLIP has been observed to enable robust zero-shot image classification when pre-training on very large Internet datasets.
We use two ViT~\cite{DBLP:conf/iclr/DosovitskiyB0WZ21} models pre-trained using CLIP: OpenAI's ViT-B/16 (pre-trained on the proprietary WebImageText (WIT) dataset) and OpenCLIP's ViT-G/14 model (pre-trained on LAION-2B~\cite{schuhmann2022laionb}).\footnote{\url{https://github.com/mlfoundations/open_clip}} 
Besides pre-training data, these models differ in their size (12 and 48 layers, respectively) and patch size (16 and 14, respectively). 
For zero-shot experiments we use these models as-is, for DP fine-tuning experiments, we use only the image encoder as a feature extractor, and on top of that, apply either a linear layer (i.e., logistic regression) or a two-layer neural network (TLNN, featuring tanh/tempered sigmoid activations~\cite{PapernotTSCE21}).
We exclusively use CLIP-based models with their respective public pre-training datasets.

%% file: sections/4_exp.tex
\section{Experiments}\label{sec:experiments}

Beyond proposing a variety of datasets and evaluation settings for benchmarking, we experimentally investigate techniques and the resulting utility obtained therein.
Some of the key questions guiding our exporation: how many of the lessons learned in DP image classification on datasets like CIFAR-10 and ImageNet transfer to the privacy-critical setting of medical images? 
How much and when does public data help for such datasets, which may be ill-represented in public data? 
And, in absolute terms, how well can we do on these datasets with DP, in various evaluation settings?

After describing our experimental setup (Section~\ref{sec:setup}), we revisit the efficacy of several ablations commonly employed in DP settings (Section~\ref{ablations}).
Finally, we make more broad conclusions about DP image classification based on our results (Section~\ref{sec:results}).
Our code is included in the code repository.


\subsection{Experimental Setup}
\label{sec:setup}
We use PyTorch~\cite{PaszkeGMLBCKLGADKYDRTCSFBC19}, and the Opacus library~\cite{YousefpourSSTPMNGBZCM21} for DP ML.
We employed the Adam optimizer~\cite{KingmaB15} across all experiments, both private and non-private, with a default learning rate of 0.001. We run our experiments at a variety of privacy levels ($\varepsilon \in [1, 3, 5, 8])$ with fixed delta values proportional to the inverse of the dataset size ($10^{-6}$ for CheXpert and $10^{-5}$ for EyePACS), as we prescribed earlier.
Batch size and total training epochs were fixed at 1024 and 20, respectively. A hyper-parameter search was performed to identify the optimal clipping norm within the range $[0.001, 0.01, 0.1, 1, 10]$. 
Following established metrics for all these datasets, we use AUC for CheXpert and EyePACS, and accuracy for CIFAR-10.
We report mean and standard deviation over three independent runs.
We used early stopping for non-private numbers due to overfitting, a phenomenon we did not observe for the DP setting due to its natural regularization properties~\cite{JungLNRSS20}.

\begin{figure*}
\centering

\begin{subfigure}{.48\linewidth}
    \includegraphics[width=\linewidth]{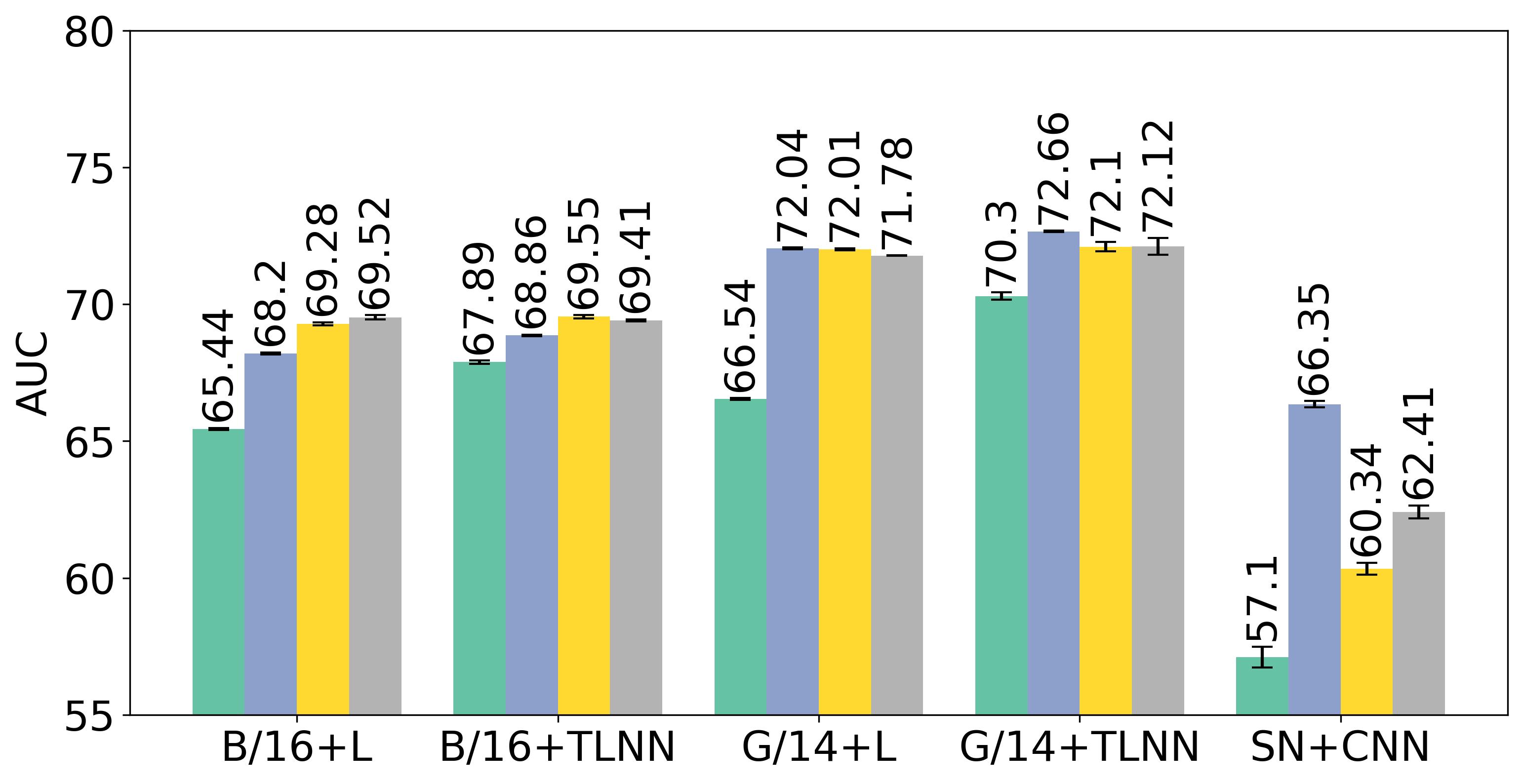}
  \caption{EyePACS}
\end{subfigure}
  \begin{subfigure}{.48\linewidth}
      \centering
  \includegraphics[width=\linewidth]{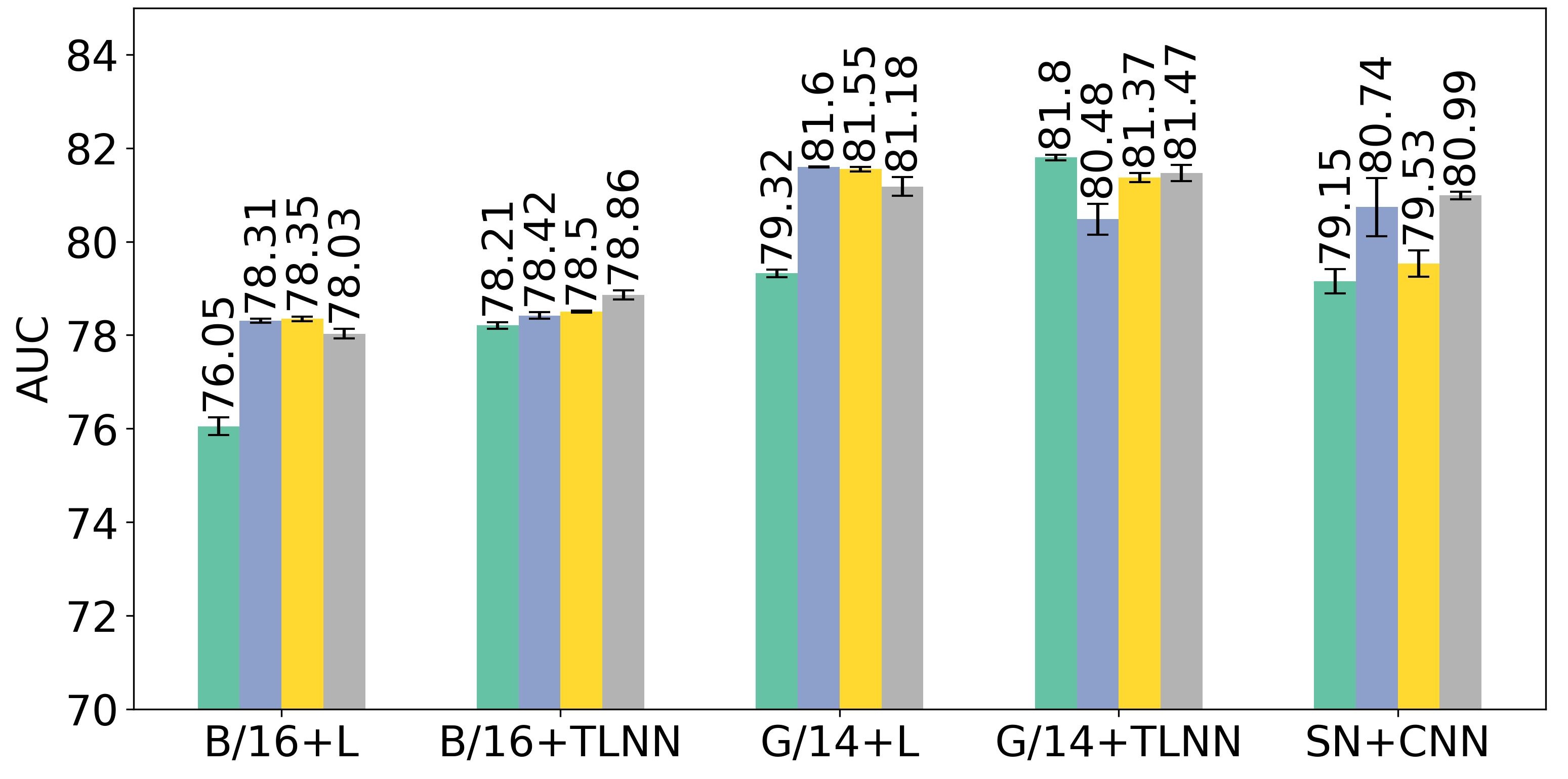}
  \caption{ CheXpert}
  \end{subfigure}
  \includegraphics[width=0.75\linewidth]{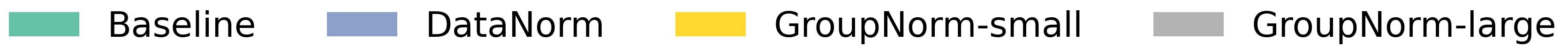}
  \caption{Normalization generally improves the final performance of all models. For CLIP-ViT models, GroupNorm small and large are groups of 8 and 16, respectively. For ScatterNets, GroupNorm small is 9 and large is 27. The choice of 27 over 81 is due to its superior performance. All experiments are done at $\varepsilon = 3$.}
  \label{fig:normalization}
  
\end{figure*}

\begin{figure*}
\centering

\begin{subfigure}{.48\linewidth}
    \includegraphics[width=\linewidth]{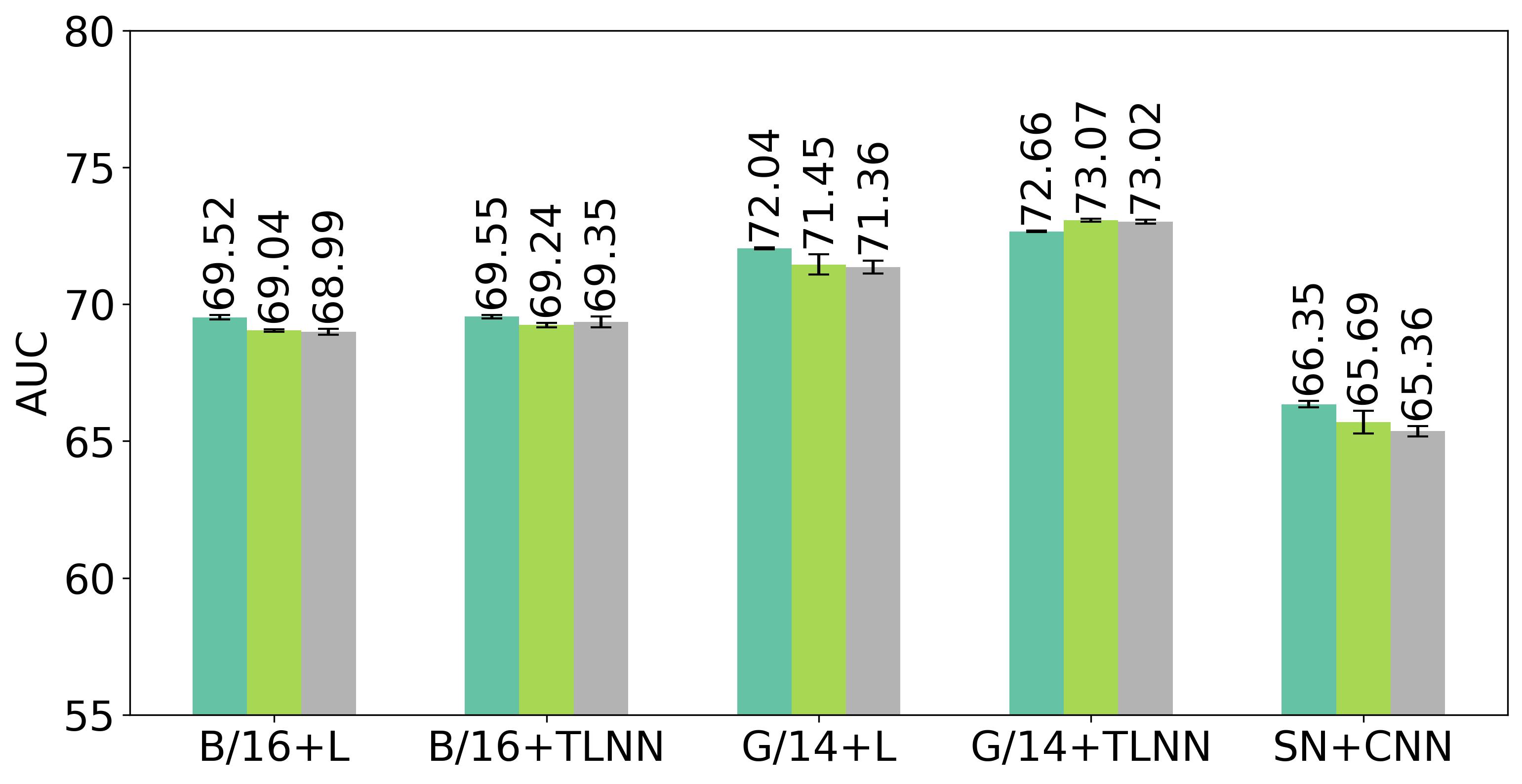}
  \caption{EyePACS}
\end{subfigure}
  \begin{subfigure}{.48\linewidth}
      \centering
  \includegraphics[width=\linewidth]{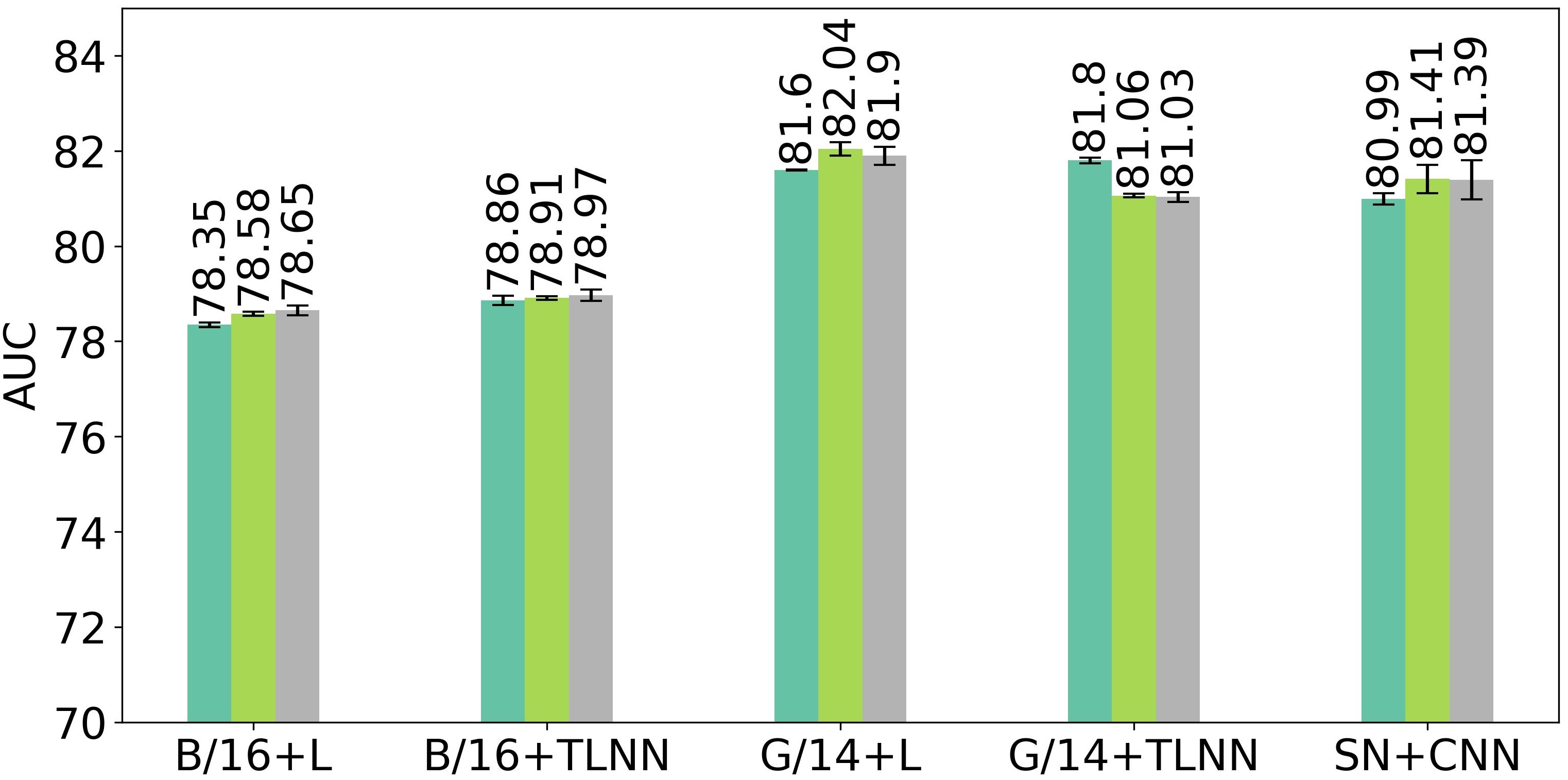}
  \caption{ CheXpert}
  \end{subfigure}
  \includegraphics[width=0.4\linewidth]{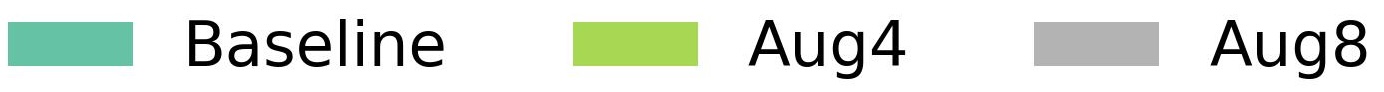}
  \caption{Augmentation multiplicity helps in general for CheXpert but not for EyePACS. We evaluate augmentation multiplicity by adding 4 and 8 augmentations of each image in the training data. All experiments are done at $\varepsilon = 3$.}
  \label{fig:augmult}
\end{figure*}
\subsection{Revisiting DP ablations}\label{ablations}
One of the most comprehensive ablation studies for DP image classification is by~\cite{DeBHSB22}.
By using group normalization, large batches, weight standardization, augmentation multiplicity, and parameter averaging, they manage to raise CIFAR-10 accuracy on a validation set from 50.8\% to an impressive 79.7\%. 
We fix $\varepsilon = 3$ and, focusing on the CLIP ViTs and ScatterNets, run the exact same ordered sequence of ablations, without carrying forward the latest technique if it does not show improved utility. Broadly speaking, while~\cite{DeBHSB22}'s techniques proved highly effective for CIFAR-10, our results reveal mixed outcomes depending on various parameters.


\paragraph{Normalization} Batch normalization~\cite{IoffeS15} is not compatible with DPSGD because it combines information across a batch, making it impossible to bound the impact of a single image in the dataset. 
Instead, prior work has shown that variants including group normalization~\cite{Wu18} and data normalization can be suitable replacements~\cite{TramerB21, DeBHSB22, BerradaDSHSSKSB23, dörmann2021noise}. 

Group normalization splits the channels of the hidden activations of an image into groups and normalizes the activations within each group. For the CLIP ViT models, with an input dimension of 512 and 1280 for the B/16 and G/14 respectively, we experiment with 8 and 16 groups. For Scatter features, with dimension (243, H/4, W/4) for RGB images and following~\cite{TramerB21}, we use 9, 27, and 81 groups. Data normalization works on data channels by normalizing using the corresponding mean and variance across the training data. Normalizing in such a way, however, incurs a privacy cost as the per-channel means and variances must be privately estimated. We use Gaussian noise with ($\sigma = 8$) to estimate these means and variances for all runs, following~\cite{TramerB21}.

In Figure~\ref{fig:normalization} we show that normalization generally improves the final performance of all models, though the most effective normalization differs across architecture and dataset. Interestingly, the four experiments where data normalization was superior involved models with larger unclipped gradients. In these cases, the optimal clipping norm chosen during hyperparameter tuning was also the highest value (10). This suggests that data normalization can effectively manage large gradient magnitudes, especially when clipping underestimates the true gradient norms. Detailed results for our experiments are given in Table~\ref{tab:scatter_normalization} and Table~\ref{tab:normalization}.


\begin{table*}[ht]
    \centering
    \caption{Studying the impact of normalization for ScatterNet + CNN, normalization consistently improves performance. Data normalization tends to outperform group normalization for EyePACS and CIFAR-10, particularly due to the large gradients of their Scatter features.}
    \resizebox{\textwidth}{!}{
    \begin{tabular}{ccccccc}
    Dataset & Model &  Baseline & DataNorm & GroupNorm9 & GroupNorm27 & GroupNorm81 \\
    \hline
    EyePACS (AUC) & SN + CNN & $57.1 \pm 0.38$ & $\textbf{66.35} \pm 0.12$ & $60.34 \pm 0.22$ & $62.41 \pm 0.24$ & $63.68 \pm 0.15$\\
    \hline
    CheXpert (AUC) & SN + CNN & $79.15 \pm 0.26$ & $80.74 \pm 0.62$ & $79.53 \pm 0.28$ & $\textbf{80.99} \pm 0.08$ & $80.72 \pm 0.35$\\
    \hline
    CIFAR-10 (Acc) & SN + CNN & $55.18 \pm 0.28$ & \boldmath{$68.29 \pm 0.17$} & $65.97 \pm 0.13$ & $66.26 \pm 0.11$ & $66.45 \pm 0.45$\\
    \end{tabular}
    }
    \label{tab:scatter_normalization}
\end{table*}

\begin{table*}[ht]
    \centering
    \caption{Normalization impact for CLIP ViT models: Normalization generally improves performance, but it also depends on the architecture and dataset. Normalizations marked in red show a drop in performance compared to the baseline.}
    \begin{tabular}{cccccc}
    Dataset & Model & Baseline &  DataNorm & GroupNorm8 & GroupNorm16 \\
    \hline
    EyePACS & B/16 + Linear & $65.44 \pm 0.04$ & $68.2 \pm 0.04$ \ & $69.28 \pm 0.06$ & \boldmath{$69.52 \pm 0.08 $}\\
    EyePACS & B/16 + TLNN & $67.89 \pm 0.06$ & $68.86 \pm 0.03$ & \boldmath{$69.55 \pm 0.06$} & $69.41 \pm 0.04$\\
    EyePACS & G/14 + Linear & $66.54 \pm 0.04$ & \boldmath{$72.04 \pm 0.03$} & $72.01 \pm 0.04$ & $71.78 \pm 0.01$\\
    EyePACS & G/14 + TLNN & $70.30 \pm 0.13$ & \boldmath{$72.66 \pm 0.03$} & $72.1 \pm 0.17$ & $72.12 \pm 0.31$\\
    EyePACS & G/14(CLIPA) + Linear & $63.88 \pm 0.08$ & \boldmath{$73.02 \pm 0.2$} & $70.7 \pm 0.06$ & $70.62 \pm 0.07$ \\
    EyePACS & G/14(CLIPA) + TLNN & $64.9 \pm 0.2$ & \boldmath{$72.9 \pm 0.1$} &  $70.87 \pm 0.25$ & $70.8 \pm 0.2$ \\
    \hline
    CheXpert & B/16 + Linear & $76.05 \pm 0.19$ & $78.31 \pm 0.04$ & \boldmath{$78.35 \pm 0.05$} & $78.03 \pm 0.1$\\
    CheXpert & B/16 + TLNN & $78.21 \pm 0.07$ & $78.42 \pm 0.07$ & $78.5 \pm 0.02$ & \boldmath{$78.86 \pm 0.1$}\\
     CheXpert & G/14 + Linear & $79.32 \pm 0.08$  & \boldmath{$81.6 \pm 0.01$} & $81.55 \pm 0.05$ & $81.18 \pm 0.2$\\
    CheXpert & G/14 + TLNN & \boldmath{$81.80 \pm 0.06$} & $80.48 \pm 0.33$ & $81.37 \pm 0.1$ & $81.47 \pm 0.17$\\
    CheXpert & G/14(CLIPA) + Linear &  $72.39 \pm 0.07$ & $76.12 \pm 0.4$ & \boldmath{$77.34 \pm 1.1$} & $77.17 \pm 0.5$\\
    CheXpert & G/14(CLIPA) + TLNN & $77.65 \pm 0.89$ &  $75.74 \pm 1.1$ & $77.38 \pm 0.3$  & \boldmath{$77.43 \pm 0.7$}\\
    \hline
    CIFAR10 & CLIP + Linear & $99.64 (93.91)$ & $99.76 (94.41)$ & $99.75 (94.43)$ & \boldmath{$99.75 (94.57)$} \\
    CIFAR10 & CLIP + TLNN & $99.69 (94.10)$ & $99.74 (94.15)$ & $99.74 (94.36)$ & \boldmath{$99.75 (94.51)$}\\
    \end{tabular}
    \label{tab:normalization}
\end{table*}

\paragraph{Larger Batch Size}
The impact of larger batch sizes in differentially private training has been observed both theoretically~\cite{bassily2020stability, TalwarTZ14} and empirically~\cite{AnilGGKM21, DeBHSB22}. In Table~\ref{tab:other_ablations}, scaling the batch size from 1024 to 4096 showed that CheXpert benefited in 80\% of experiments, while EyePACS did not. This disparity is likely due to CheXpert having a training set six times larger than EyePACS, resulting in fewer model update steps for EyePACS and potential underfitting with a fixed number of epochs. We further observed that increasing the number of epochs showed a positive impact of larger batch sizes on EyePACS when using the ScatterNet model.

\paragraph{Weight Standardization} 
We experiment with weight standardization (WS) on the Scatternet + CNN model as it applies to only convolution layers. From our results in Table~\ref{tab:other_ablations}, we observe that weight standardization does not help with EyePACS but helps with CheXpert and CIFAR-10. As alluded by prior work~\cite{brock2021, DeBHSB22}, we also observe a positive correlation of group normalization with WS. However, due to a limited number of experiments, we do not have strong evidence either way.

\begin{figure*}
\centering

\begin{subfigure}{.48\linewidth}
    \includegraphics[width=\linewidth]{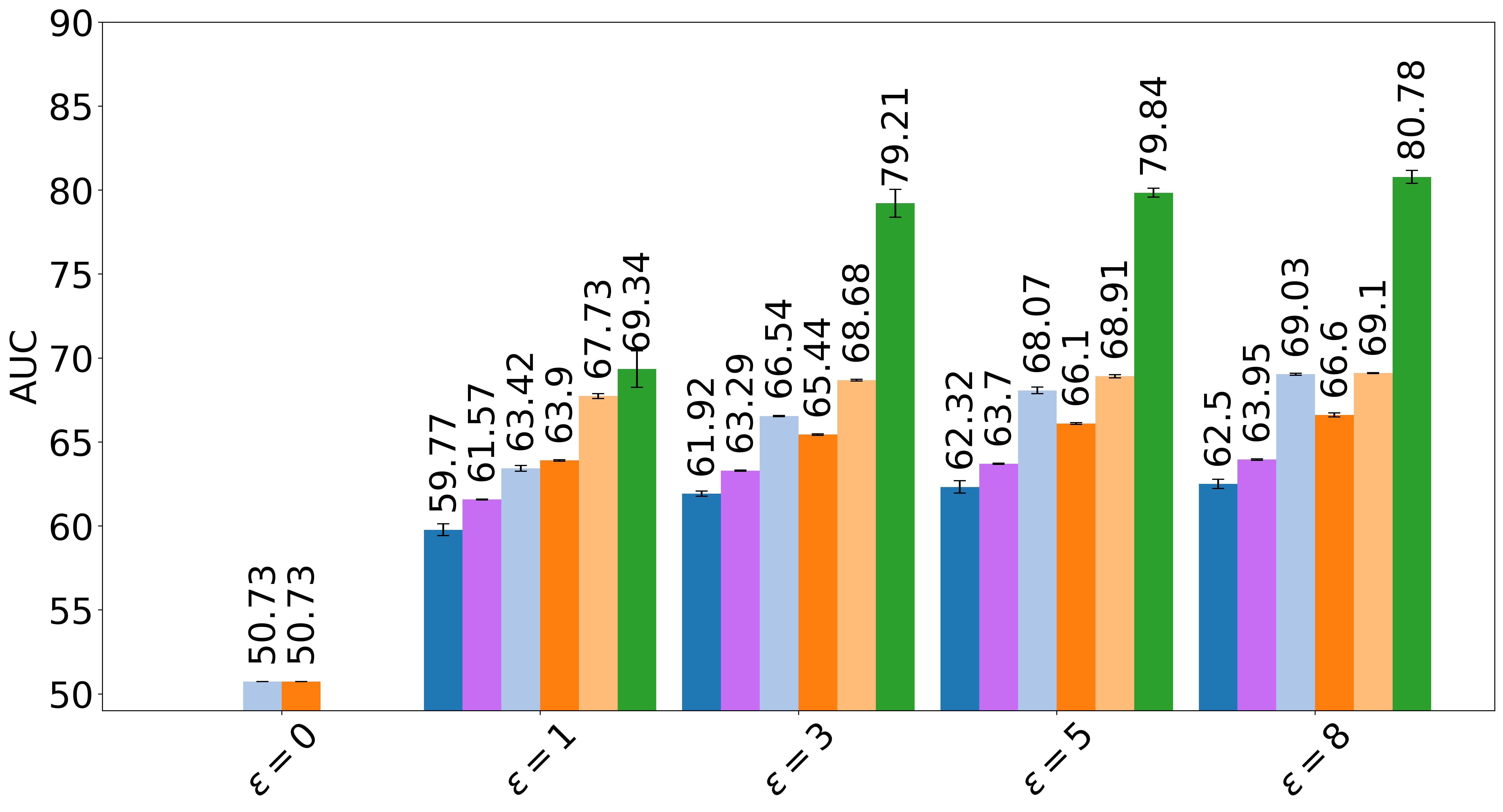}
  \caption{EyePACS}
\end{subfigure}
  \begin{subfigure}{.48\linewidth}
      \centering
  \includegraphics[width=\linewidth]{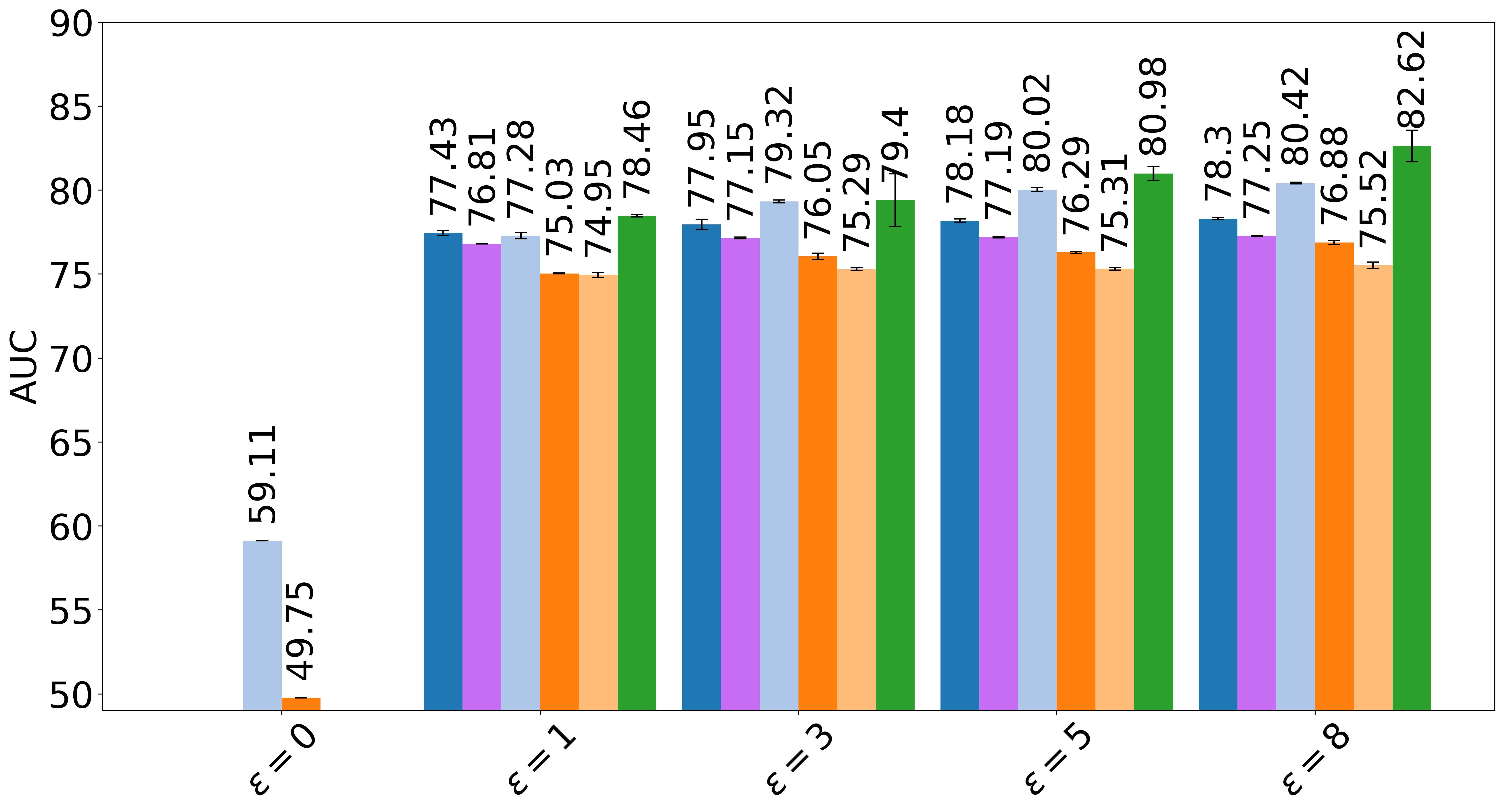}
  \caption{ CheXpert}
  \end{subfigure}
  \includegraphics[width=0.75\linewidth]{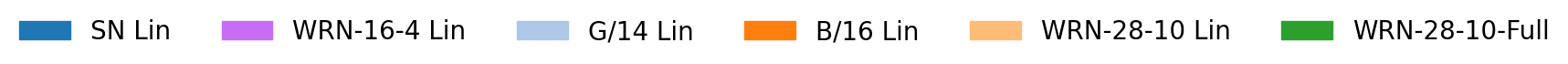}
  \caption{Pre-training datasets have different impacts: Wide-ResNet, pretrained on ImageNet, performs best on EyePACS, while ViT-G/14 with linear probe surpasses Wide-ResNet 28-10 linear probe across all $\varepsilon$ values on CheXpert. Furthermore, ViT-G/14 achieves near-random performance on EyePACS in zero-shot settings but attains a non-trivial 59.11\% AUC on CheXpert.}
  \label{fig:pretraining}
  
\end{figure*}

\begin{figure*}\label{fig:norm}
\centering

\begin{subfigure}{.48\linewidth}
    \includegraphics[width=\linewidth]{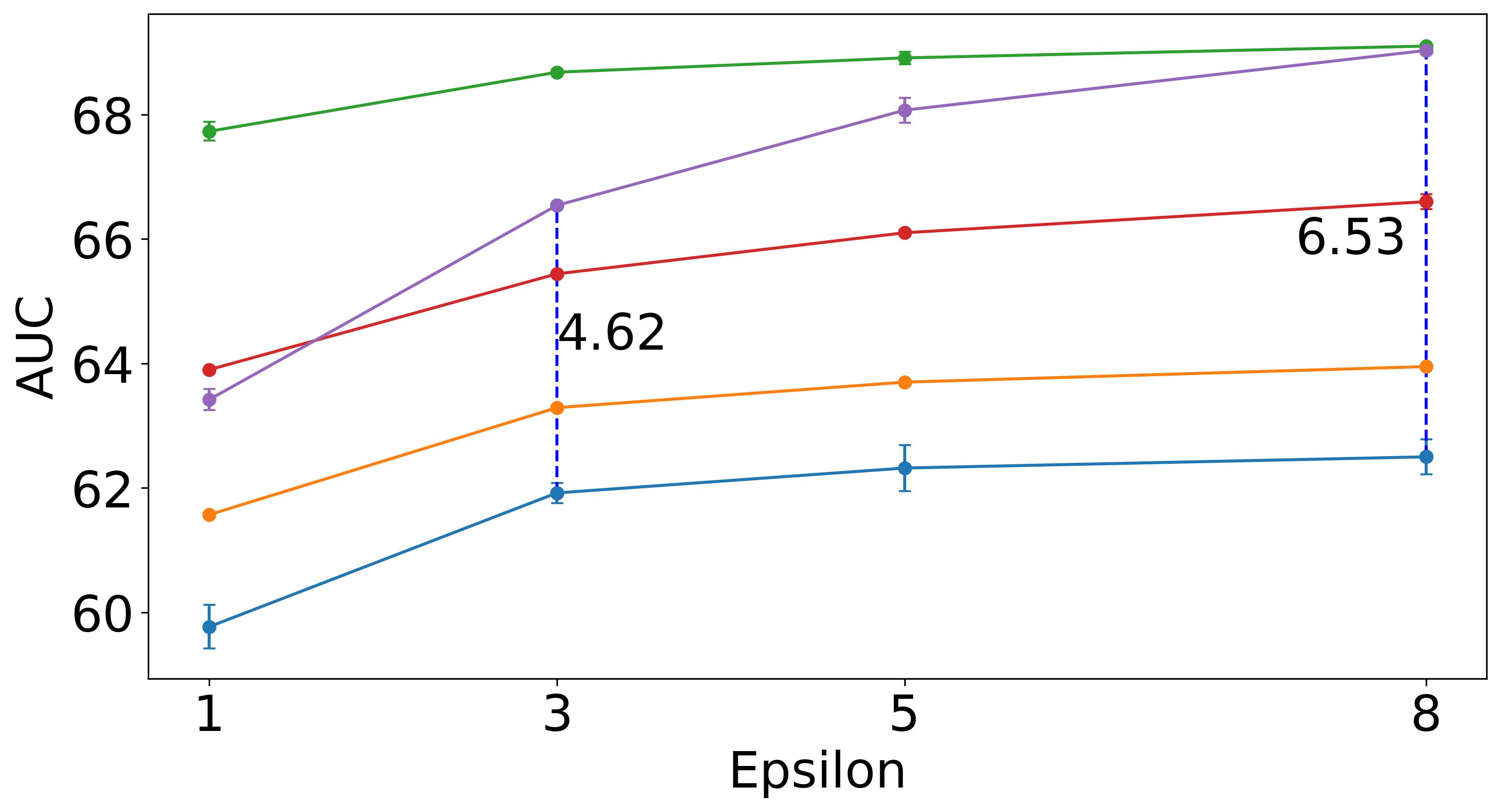}
  \caption{EyePACS}
\end{subfigure}
  \begin{subfigure}{.48\linewidth}
      \centering
  \includegraphics[width=\linewidth]{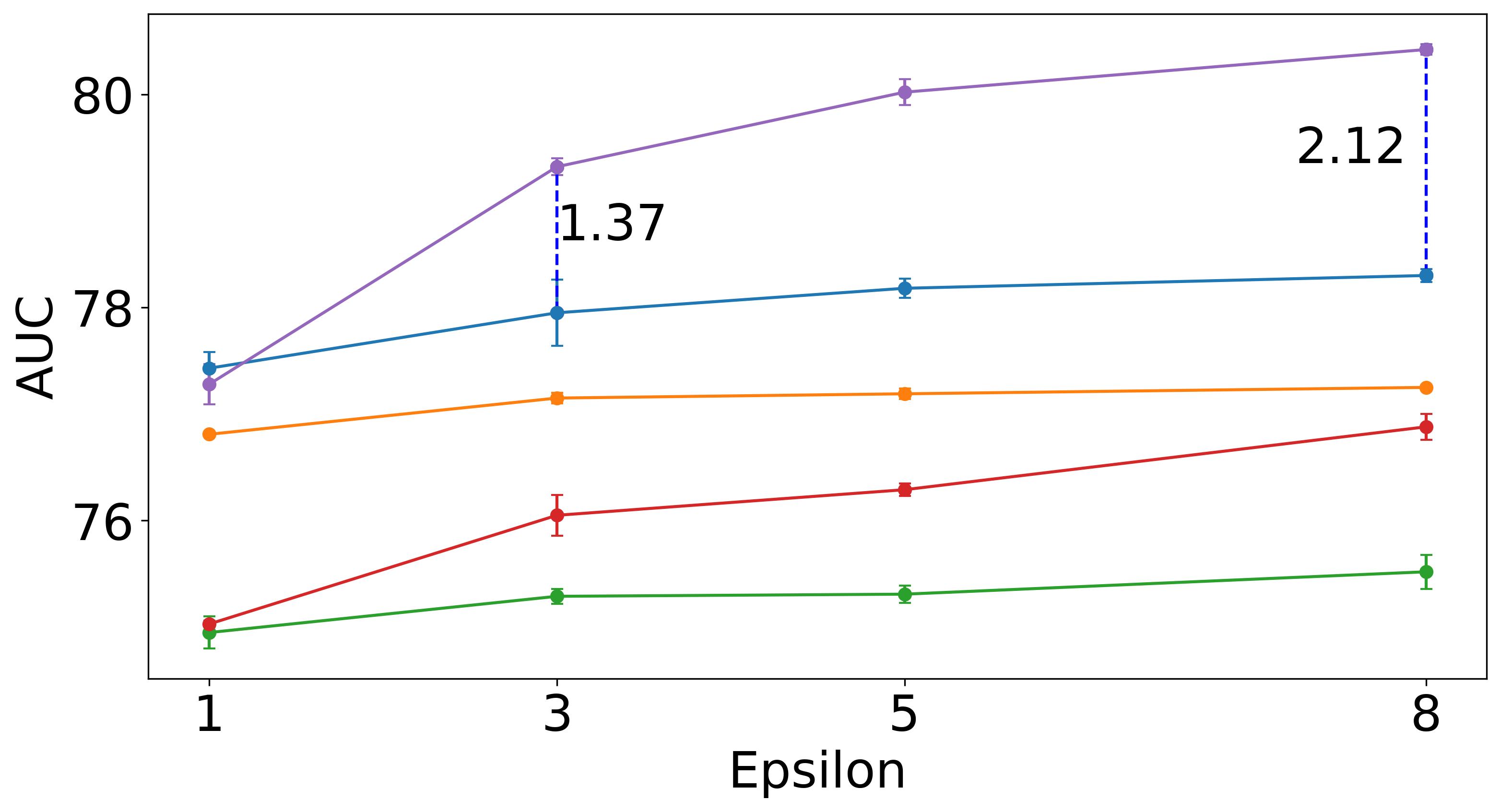}
  \caption{ CheXpert}
  \end{subfigure}
  \includegraphics[width=0.75\linewidth]{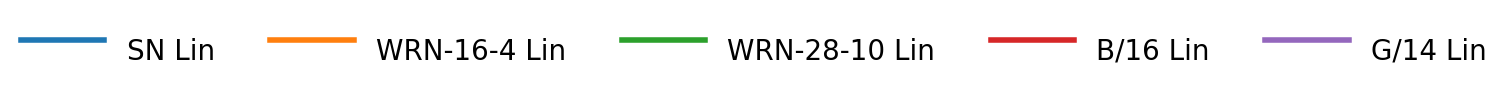}
  \caption{Pre-training public data is more beneficial with higher $\varepsilon$ values. For CheXpert, ScatterNet performs better at smaller $\varepsilon$ values, while pretrained models show marginal improvements at larger $\varepsilon$ values. Similarly, for EyePACS, CLIP ViT-G/14 linear performs better as $\varepsilon$ value increases.}
  \label{fig:larger_epsilon}
  
\end{figure*}

\begin{table*}[ht]
    \centering
    \caption{Studying all ablations together, we observe an almost consistent improvement in performance for CIFAR-10, whereas this pattern is not observed with the other datasets.}\label{tab:other_ablations}
    \resizebox{\textwidth}{!}{
    \begin{tabular}{ccccccc}
    Dataset & Model & Best Normalization & +Larger Batch & +WS & +Best Augmult & +EMA \\
    \hline
    EyePACS & SN + CNN & $66.35 \pm 0.12$ & $66.24 \pm 0.23$ & $65.81 \pm 0.06$ & $65.69 \pm 0.42$ & \boldmath{$66.65 \pm 0.16$}\\
    EyePACS & B/16 + L & \boldmath{$69.52 \pm 0.08$} & $68.95 \pm 0.24$ & - & $69.04 \pm 0.05$ & $69.09 \pm 0.12$\\
    EyePACS & B/16 + TLNN & \boldmath{$69.55 \pm 0.06$} & $69.16 \pm 0.28$ & - & $69.24 \pm 0.08$& $69.3 \pm 0.21$\\
    EyePACS & G/14 + L & \boldmath{$72.04 \pm 0.03$} & $71.59 \pm 0.24$ & - & $71.45 \pm 0.37$ & $71.29 \pm 0.17$\\
    EyePACS & G/14 + TLNN & $72.66 \pm 0.03$ & \boldmath{$73.07 \pm 0.06$} & - & $73.07 \pm 0.05$& $73 \pm 0.03$\\
    EyePACS & CLIPA + L & \boldmath{$73.02 \pm 0.2$} & $68.5 \pm 0.07$ & - & $68.37 \pm 0.05$ & $70.94 \pm 0$ \\
    EyePACS & CLIPA + TLNN & $72.09 \pm 0.1$ & \boldmath{$72.17 \pm 0.1$} & - & $72.04 \pm 1.6$ & $71.9 \pm 0.07$ \\
    \hline
    CheXpert & SN + CNN & $80.99 \pm 0.08$ & $81.54 \pm 0.34$ & $82.11 \pm 0.29$ & $81.41 \pm 0.3$ & \boldmath{$82.24 \pm 0.22$}\\
    CheXpert & B/16 + L & $78.35 \pm 0.05$ & $78.42 \pm 0.07$ & - & \boldmath{$78.65 \pm 0.1$} & \boldmath{$78.65 \pm 0.04$}\\
    CheXpert & B/16 + TLNN & $78.86 \pm 0.1$ & $78.86 \pm 0.1$ & - & $78.97 \pm 0.12$ & \boldmath{$79.01 \pm 0.1$}\\
    CheXpert & G/14 + L & $81.6 \pm 0.01$ & $81.76 \pm 0.05$ & - & \boldmath{$82.04 \pm 0.14$} & $82.01 \pm 0.05$\\
    CheXpert & G/14 + TLNN & $81.8 \pm 0.06$ & $81.04 \pm 0.16$ & - & $81.06 \pm 0.34$ & $81.14 \pm 0.26$ \\
    CheXpert & CLIPA + L &  $77.34 \pm 1.1$ & $80.17 \pm 0.08$ & - & \boldmath{$80.38 \pm 1.5$} & $80.34 \pm 0.15$\\
    CheXpert & CLIPA + TLNN & $77.43 \pm 0.7$ & $80.4 \pm 0.2$ & - & \boldmath{$80.75 \pm 0.18$} & $80.52 \pm 0.02$\\
    \hline
    CIFAR10 & B/16 + L & $99.75(94.57)$ & $99.74 (94.49)$ & - & \boldmath{$99.77(94.76)$} & $99.76 (94.67)$\\
    CIFAR10 & B/16 + TLNN & $99.75(94.51)$ & $99.76(94.55)$ & - & \boldmath{$99.79(94.81)$} & $99.78(94.76)$\\
    CIFAR-10 & SN + CNN & $68.29 \pm 0.17$ & $66.47 \pm 0.31$ & $68.96 \pm 0.26$& \boldmath{$69.16 \pm 0.08$} & $68.07 \pm 0.24$ \\ 
    \end{tabular}
    }
    \par
\end{table*}

\paragraph{Augmentation Multiplicity}

We apply a sequence of augmentations to our benchmark datasets: reflect padding, random cropping, and random horizontal flipping. While \cite{DeBHSB22} recommend 16 augmentations per image, due to computational constraints with large datasets, we use 4 and 8 augmentations. As shown in Figure~\ref{fig:augmult}, contrary to \cite{DeBHSB22}'s findings, augmentation multiplicity (augmult) does not consistently yield positive effects. Except for one experiment, (ViT-G/14+TLNN), augmentations generally benefit CheXpert but not EyePACS. Future work may explore the effectiveness of dataset-specific augmentations, which could potentially yield more beneficial results. We show detailed experiment results in Table~\ref{tab:augmult}.

\begin{table*}[ht]
    \centering
    \caption{Studying the impact of augmentation multiplicity, we find that it consistently improves performance for CIFAR-10. However, looking at EyePACS and CheXpert, we observe inconsistent behavior, except that it generally seems to reduce performance with EyePACS. For the third column, we take the best result from Table~\ref{tab:other_ablations} after best normalization, larger batch size, and weight standardization. 
}
    \begin{tabular}{ccccc}
    Dataset & Model &  Norm + Larger BS + WS  & +Augmult(4) & +Augmult(8) \\
    \hline
    EyePACS & SN + CNN & \boldmath{$66.35 \pm 0.12$} & $65.69 \pm 0.42$ & $65.36 \pm 0.19$ \\
    EyePACS & B/16 + L & \boldmath{$69.52 \pm 0.08$} & $69.04 \pm 0.05$ & $68.99 \pm 0.11$ \\
    EyePACS & B/16 + TLNN & \boldmath{$69.55 \pm 0.06$} & $69.24 \pm 0.08$ & $69.35 \pm 0.2$ \\
    EyePACS & G/14 + L & $72.04 \pm 0.03$ & $71.45 \pm 0.37$ & $71.36 \pm 0.24$ \\
    EyePACS & G/14 + TLNN & $73.07 \pm 0.05$ & \boldmath{$73.07 \pm 0.05$} & $73.02 \pm 0.07$ \\
    EyePACS & G/14 (CLIPA)+ L & $73.02 \pm 0.2$ & $68.37 \pm 0.05$ & $68.14 \pm 1.2$ \\
    EyePACS & G/14 (CLIPA)+ TLNN & $72.9 \pm 0.1$ & $72.04 \pm 1.6$ & $71.98 \pm 0.3$ \\
    \hline
    CheXpert & SN + CNN & $82.11 \pm 0.29$ & $81.41 \pm 0.3$ & $81.39 \pm 0.41$ \\
    CheXpert & B/16 + L & $78.42 \pm 0.07$ & $78.58 \pm 0.04$ & \boldmath{$78.65 \pm 0.1$} \\
    CheXpert & B/16 + TLNN & $78.86 \pm 0.1$ & $78.91 \pm 0.04$ & \boldmath{$78.97 \pm 0.12$} \\
    CheXpert & G/14 + L & $81.76 \pm 0.05$ & \boldmath{$82.04 \pm 0.14$} & $81.9 \pm 0.19$ \\
    CheXpert & G/14 + TLNN & $81.8 \pm 0.06$ & $81.06 \pm 0.04$  & $81.03 \pm 0.1$ \\
    CheXpert & G/14 (CLIPA)+ L & $77.34 \pm 1.1$ & $80.34 \pm 0.2$ & \boldmath{$80.38 \pm 1.5$} \\
    CheXpert & G/14 (CLIPA)+ TLNN & $77.43 \pm 0.7$ & $80.5 \pm 0.1$ & \boldmath{$80.75 \pm 0.18$} \\
    \hline
    CIFAR-10 (ACC)& SN + CNN & $68.96 \pm 0.26$ & $69.07 \pm 0.2$ & $\textbf{69.16} \pm 0.08$ \\
    CIFAR-10 (ACC)& B/16 + L & $99.75 (94.57)$ & $99.76 (94.68)$ & \boldmath{$99.77	(94.76)$} \\
    CIFAR-10 (ACC)& B/16 + TLNN & $99.76 (94.55)$ & $99.78 (94.75)$ & \boldmath{$99.79 (94.81)$} \\
    
    \hline 
    \end{tabular}
    \label{tab:augmult}
\end{table*}

\paragraph{Parameter Averaging} 
The final ablation that~\cite{DeBHSB22} suggests is the exponential moving average (EMA)\cite{PolyakEMA} of all the parameters in the model. In Table~\ref{tab:other_ablations}, we notice that EMA occasionally improves performance, which contradicts the findings of \cite{DeBHSB22} that it consistently enhances results across all experiments.

%% file: sections/5_results.tex
\subsection{Experimental findings}\label{sec:results} 

We highlight some findings from our experimental results.

\paragraph{Different pre-training datasets offer varying degrees of improvement depending on the private data}
We compare representative models publicly pre-trained on a variety of datasets on both CheXpert and EyePACS. 
Results are displayed in Figure~\ref{fig:pretraining}. For the case of no pre-training data, we choose ScatterNet+Linear, due to its consistently superior utility compared to Wide-ResNet trained from scratch, particularly for high privacy (i.e., low $\varepsilon$) settings.

On the other end of the spectrum, when we allow  large-scale public pre-trarining, the CLIP ViT models provide a good indication of zero-shot performance (i.e., $\varepsilon = 0$).

When analyzing CheXpert, ViT-B/16 performs close to random in the zero-shot setting, whereas ViT-G/14 achieves an AUC of 59.11\%, moderately better than random. Moving from ScatterNet+linear to Wide-ResNet+Linear, there is a noticeable decrease in AUC, yet ViT-G/14 consistently outperforms across various $\varepsilon$ values, indicating that ViT-G/14 is a better fit for CheXpert. Notably, at $\varepsilon = 8$, Wide-ResNet with full fine-tuning exceeds the performance of ViT-G/14. However, considering that Wide-ResNet is fully fine-tuned while ViT-G/14 is not, this doesn't necessarily make Wide-ResNet better suited for CheXpert. Nevertheless, for smaller $\varepsilon$, it is clear that ViT-G/14 is the superior model with only linear fine-tuning.


Looking at Figure \ref{fig:pretraining} for EyePACS, both CLIP ViT models show random performance in the zero-shot setting, indicating no improvement from pretraining. Conversely, Wide-ResNet linear exhibits a significant performance boost when transitioning from ScatterNet linear to Wide-ResNet linear, maintaining its superiority across all $\varepsilon$ values. Although we notice that as we move toward less private regimes, the power of pre-trained ViT-G/14 becomes more evident, particularly from $\varepsilon=1$ to $\varepsilon=3$, approaching the performance of Wide-ResNet linear. However, there remains a substantial gap between fully fine-tuned Wide-ResNet and the other models, unlike CheXpert, suggesting that Wide-ResNet is better suited for EyePACS.

\paragraph{Public pre-training data helps more with higher $\varepsilon$ values}

\begin{table*}[ht]
    \centering
    \begin{center}
    \caption{Test AUC for EyePACS at different epsilons. Baselines include ScatterNet (SN), WideResNet (WRN) and CLIP models on datasets with different public data pre-training. The SOTA is due to \cite{Voets_2019}.}
    \resizebox{\textwidth}{!}{
        \begin{tabular}{lcccccccc}
            \hline \multirow{2}{*}{\text { Public data }} & \multirow{2}{*} {\text { Model }} & \multicolumn{5}{c}{\text { Test AUC (\%) }}\\
            \cline { 3 - 7 } &  & $\varepsilon=1$ & $\varepsilon=3$ & $\varepsilon=5$ & $\varepsilon=8$ & $\varepsilon=\infty$\\
            
            \hline 
            
            \text{ None} & \text {SN + L} & $59.77\pm0.35$ & $61.92\pm0.16$ & $62.32\pm0.37$ & $62.5\pm0.28$ & $69.70\pm0.11$ \\
            \text { None} & \text {SN + CNN} & $63.73\pm0.11$ & $66.36\pm0.17$ & $66.59\pm0.43$ & $67.37\pm0.27$ & $69.28\pm0.20$\\ 
            \text { None} & \text {WRN-16-4+L} & $61.57\pm0.02$ & $63.29\pm0.04$ & $63.7\pm0.04$ & $63.95\pm0.04$ & $67.74\pm0.01$\\
            \text { None} & \text {WRN (Scratch)} & $55.45\pm0.18$ & $56.53\pm0.08$ & $57.14\pm0.34$ & $57.65\pm0.22$ & $61.97\pm0.09$\\
            \text { IN-32  } & \text {WRN +  Linear} & $67.73\pm0.15$ & $68.68\pm0.05$ & $68.91\pm0.10$ & $69.10\pm0.03$ & $73.21\pm0.09$\\
            \text { IN-32 } & \text {WRN (Full)} & $69.34\pm1.09$ & $79.21\pm0.83$ & $79.84\pm0.27$ & $80.78\pm0.38$ & $83.61\pm0.03$\\
            \text { WIT } & \text {B/16 + Linear} & $63.9 \pm 0.04$ & $65.44 \pm 0.04$ & $66.1 \pm 0.05$& $66.6 \pm 0.12$ & $69.93\pm 0.01$ \\
            \text { WIT } & \text {B/16 + TLNN} & $65.12 \pm 0.04$ & $67.89 \pm 0.06$ &  $69.22 \pm 0.1$ & $69.84 \pm 0.02$ & $70.54 \pm 0.01$\\
            \text { LAION } & \text {G/14 + Linear} & $63.42 \pm 0.17$ & $66.54 \pm 0.04$ & $68.07 \pm 0.2$ & $69.03 \pm 0.06$ & $69.88 \pm 0.2$ \\
            \text { LAION  } & \text {G/14 + TLNN} & $65.47 \pm 0.02$ & $70.30 \pm 0.13$ &  $71.74 \pm 0.3$ & $72.3 \pm 0.19$ & $73.36 \pm 0.15$\\
            \text { DataComp1B } & \text {G/14(CLIPA) + Linear} & $63.42 \pm 0.06$ & $63.88 \pm 0.08$ & $63.8 \pm 0.12$ & $64.33 \pm 0.32$ & $70.87 \pm 0.01$ \\
            \text { DataComp1B  } & \text {G/14(CLIPA) + TLNN} & $64.41 \pm 1.00$ & $64.9 \pm 0.20$ &  $65.07 \pm 0.26$ & $65.67 \pm 0.08$ & $75.42 \pm 0.08$\\
            \text { IN-1K } & \text {SOTA} & - & - & -  & - & $95.1$ \\
            \hline
        \end{tabular}\label{tab:eyepacs_full}
    }
    \end{center}
\end{table*}

\begin{table*}[ht]
    \centering
    \begin{center}
    \caption{Test AUC for CheXpert at different epsilons. Baselines include ScatterNet (SN), WideResNet (WRN) and CLIP models on datasets with different public data pre-training. The SOTA is from \cite{BerradaDSHSSKSB23}(Private) and \cite{yuan2021large}(Non-Private).}
    \resizebox{\textwidth}{!}{
        \begin{tabular}{lcccccccc}
            \hline \multirow{2}{*}{\text { Public data }} & \multirow{2}{*} {\text { Model }} & \multicolumn{5}{c}{\text { Test AUC (\%) }}\\
            \cline { 3 - 7 } &  & $\varepsilon=1$ & $\varepsilon=3$ & $\varepsilon=5$ & $\varepsilon=8$ & $\varepsilon=\infty$\\
            \hline \text { None} & \text { SN + CNN } & $78.16 \pm 0.22$ & $79.15 \pm 0.26$ & $79.16 \pm 0.18 $ & $79.68 \pm 0.04$ & $80.65 \pm 0.12$\\
            \text { None } & \text { SN + Linear } & $77.43 \pm 0.15$ & $77.95 \pm 0.31$ & $78.18 \pm 0.09$ & $78.30 \pm 0.06$ & $78.94 \pm 0.13 $\\
            \text { None} & \text { WRN-16-4+L} & $76.81\pm0.02$ & $77.15\pm0.05$ & $77.19\pm0.05$ & $77.25\pm0.01$ & $77.49\pm0.03$\\
            \text { None } & \text {WRN (Scratch)} & $76.9\pm0.02$ & $77.8\pm0.04$ & $77.89\pm0.05$ & $78.68\pm0.10$ & $87.31\pm0.07$\\
            \text { IN-32 } & \text {WRN + Linear} & $74.95\pm0.15$ & $75.29\pm0.07$ & $75.31\pm0.08$ & $75.52\pm0.16$ & $75.91\pm0.08$\\            
            \text { IN-32 } & \text { WRN (Full)} & $78.46\pm0.07$ & $79.40\pm1.57$ & $80.98\pm0.42$ & $82.62\pm0.94$ & $87.62\pm0.09$\\
            \text { WIT } & \text { B/16 + Linear} &  $75.03 \pm 0.03$ & $76.05 \pm 0.19$ & $76.29 \pm 0.06$ & $76.88 \pm 0.12$ & $76.89 \pm 0.01$ \\
            \text { WIT } & \text { B/16 + TLNN } & $77.28 \pm 0.19$ & $78.21 \pm 0.07$ & $78.33 \pm 0.04$ & $78.54\pm 0.06$ & $78.56 \pm 0.01$\\
            \text { LAION } & \text {  G/14 + Linear} & $77.28 \pm 0.19$ & $79.32 \pm 0.08$ & $80.02 \pm 0.12$ & $80.42\pm 0.05$ & $80.48 \pm 0.02$\\
            \text { LAION } & \text { G/14 + TLNN } & $80.63 \pm 0.4$ & $81.80 \pm 0.06$ & $82.25 \pm 0.02$ & $82.27 \pm 0.0.4$ & $82.28 \pm 0.01$\\
            \text { DataComp1B } & \text {G/14(CLIPA) + Linear} & $71.45 \pm 0.27$ & $72.39 \pm 0.07$ & $72.98 \pm 0.38$ & $72.37 \pm 0.41$ & $78.35 \pm 1.3 $\\
            \text { DataComp1B  } & \text {G/14(CLIPA) + TLNN} & $77.3 \pm 0.60$ & $77.65 \pm 0.89$ &  $77.67 \pm 0.25$ & $77.51 \pm 0.85$ & $80.62 \pm 0.06$\\
            \text { IN-21K } & \text { SOTA} & $86.3$ & - & - & $89.2$ & - \\
            \text { IN-1K } & \text { SOTA} & - & - & - & - & $93$ \tablefootnote{The authors use an ensemble of five models.}\\
            \hline
        \end{tabular}\label{tab:chexpert_full}
    }
    \end{center}
\end{table*}

We compare feature generation methods in Figure \ref{fig:larger_epsilon} since, in all cases, there is a linear classifier on top of diverse feature extractors. On CheXpert, linear fine-tuning with ScatterNet shows the best performance at $\varepsilon=1$. However, as $\varepsilon$ increases, pretrained models, especially ViT-G/14, begin to outperform other methods significantly. While full fine-tuning of CLIP has not been explored, a direct comparison of features shows ViT-G/14's superiority when $\varepsilon$ is sufficiently large. As $\varepsilon$ value increases further, ViT-G/14's performance improves notably, highlighting its strong pretrained performance under less stringent privacy constraints.

When comparing the best performance on CheXpert across our proposed methods, ScatterNet achieves superior results compared to CLIP ViT models and Wide-ResNet on $\varepsilon = 3$, as shown in Figure \ref{fig:larger_epsilon}. However, as $\varepsilon$ values increase, pretrained models begin to perform better, and the performance gap between ScatterNet and the other models widens.

For EyePACS, we don't see the same pattern, likely because EyePACS is a much smaller dataset (about one-sixth the size) and Scatter features have high dimensionality, making it hard to balance this complexity with private training. We use ScatterNet linear as the baseline for the no pretraining regime and compare it to other architectures' linear fine-tuning for a fair comparison.

As illustrated in Figure \ref{fig:larger_epsilon}, increasing the $\varepsilon$ value amplifies ViT-G/14's performance advantage over the ScatterNet baseline, widening the gap. However, we do not observe any significant changes in ViT-B/16 and Wide-ResNet linear. ViT-B/16 appears to perform poorly regardless of privacy settings. On the other hand, Wide-ResNet linear consistently maintains a significant gap between its linear model and ScatterNet. This can be explained by the fact that Wide-ResNet linear can already achieve high AUC in the $\varepsilon=1$ case, leaving little room for improvement.

The fact that Wide-ResNet maintains its advantage from the start is not surprising, given that as discussed earlier, the pre-trained model seems to help with EyePACS the most. However, ViT-G/14's performance improves more as the $\varepsilon$ value increases. The detailed numbers for this experiment are provided in Table~\ref{tab:eyepacs_full} and Table~\ref{tab:chexpert_full}.

\paragraph{Progress on CIFAR10 does not translate to progress on benchmark datasets}

Looking at Figure \ref{fig:pretraining}, we notice that ViT-G/14 achieves an astonishing 99.75\% zero-shot accuracy on CIFAR-10. In stark contrast, the same model's zero-shot performance on CheXpert and EyePACS is significantly lower, with AUC scores of 59.11\% and 50.73\%, respectively—the latter essentially equating to random guessing. Additionally, Wide-ResNet achieves 94.7\% accuracy on CIFAR-10 at $\varepsilon=1$, yet only 78.52\% and 71.00\% AUC on CheXpert and EyePACS, respectively.

Upon reviewing the ablation experiments in section \ref{ablations}, it becomes evident that the techniques beneficial for CIFAR-10 do not necessarily yield similar advantages for EyePACS and CheXpert datasets. The patterns observed in CIFAR-10 did not replicate in these medical image datasets, and notably, performance on CheXpert and EyePACS showed inconsistency.

Additionally, we observe that incorporating synthetic data as demonstrated by Tang et al.~\cite{tang2023differentiallyprivateimageclassification}, leads to SOTA performance on CIFAR-10 without pretraining. However, in our experiments, ScatterNet outperforms~\cite{tang2023differentiallyprivateimageclassification}'s approach on CheXpert, whereas on EyePACS, Tang et al. achieve better results.



%% file: sections/6_relatedfuture.tex
\section{Related Work}
Several works have evaluated the privacy-utility tradeoffs for DPML algorithms~\cite{zhao2020not, jayaraman2019evaluating, jarin2022dp}. Jayaraman et al.~\cite{jayaraman2019evaluating} explored the impact of various variants of DP for ML algorithms. They explored the privacy leakage concerning the privacy parameter $\epsilon$ for the same algorithm. The work of Zhao et al.~\cite{zhao2020not} and Jarin et al.~\cite{jarin2022dp} similarly study the privacy-utility tradeoffs for different DP ML algorithms and evaluate them against membership inference attacks. There have also been some attempts at benchmarking DP algorithms~\cite{tao2022benchmarkingdifferentiallyprivatesynthetic, wei2023dpmlbench, gong2025dpimagebenchunifiedbenchmarkdifferentially}. Tao et al.~\cite{tao2022benchmarkingdifferentiallyprivatesynthetic} and Gong et al.~\cite{gong2025dpimagebenchunifiedbenchmarkdifferentially} benchmark different synthetic data generation algorithms for tabular data and image data respectively. The work of Wei et al.~\cite{wei2023dpmlbench} is closest to our work, where they benchmark different DPML algorithms on standard ML datasets such as MNIST/CIFAR-10 and comment on the effects of improvements made in DPML literature. In our work, we take a different stance than them and propose a new benchmark based on privacy-critical medical datasets. Compared to their work, we also experimented with more established architectures based on various techniques, such as Scatternets and CLIP-based models. 

\section{Future Work}

While our work focused on image classification, future research should explore benchmarks in other areas such as Natural Language Understanding and Generation. In addition, to ensure fair comparisons, future work could investigate the use of more advanced model architectures. For instance, experiments using the NFNet-F7~\cite{brock2021highperformancelargescaleimagerecognition} model pre-trained on ImageNet-1K could be compared with our Wide-ResNet experiments.

Future research should also extend to a wider range of datasets, both within and beyond the medical domain. This exploration will help in understanding the generalizability of DP ML techniques and identifying domain-specific challenges.

The continued maintenance and updating of the leaderboard we have established will be crucial for tracking long-term progress in the field and identifying emerging trends or breakthroughs. This ongoing effort will provide valuable insights into the evolution of DP ML techniques over time.

%% file: sections/7_conclusion.tex
\section{Conclusion}\label{sec:conclusion}
We suggest a number of standardized settings for benchmarking DP image classification, particularly with a focus on privacy-critical domains such as medical images.
We also provide a leaderboard to help track progress on image classification benchmarks.
In our experimental investigation, we find that several of the techniques which have enjoyed great success for DP ML are \emph{not} universally effective across datasets and architectures, and furthermore that progress on standard benchmarks like CIFAR-10 do \emph{not} transfer to medical images. 
Of course, it is hard and rare to design universally effective techniques. 
Indeed, our experiments are for a limited number of datasets and a limited number of architectures, so it is impossible to make a conclusion broad enough to encompass the entire field of DP image classification. 
However, it is clear that present work leaves the door open for new ideas and techniques that push the envelope on private image classification in these settings.

%% file: biblio.bib
@inproceedings{jayaraman2019evaluating,
  title={Evaluating differentially private machine learning in practice},
  author={Jayaraman, Bargav and Evans, David},
  booktitle={28th USENIX Security Symposium (USENIX Security 19)},
  pages={1895--1912},
  year={2019}
}

@inproceedings{zhao2020not,
  title={Not one but many tradeoffs: Privacy vs. utility in differentially private machine learning},
  author={Zhao, Benjamin Zi Hao and Kaafar, Mohamed Ali and Kourtellis, Nicolas},
  booktitle={Proceedings of the 2020 ACM SIGSAC Conference on Cloud Computing Security Workshop},
  pages={15--26},
  year={2020}
}

@misc{gong2025dpimagebenchunifiedbenchmarkdifferentially,
      title={DPImageBench: A Unified Benchmark for Differentially Private Image Synthesis}, 
      author={Chen Gong and Kecen Li and Zinan Lin and Tianhao Wang},
      year={2025},
      eprint={2503.14681},
      archivePrefix={arXiv},
      primaryClass={cs.CR},
      url={https://arxiv.org/abs/2503.14681}, 
}

@inproceedings{jarin2022dp,
  title={Dp-util: comprehensive utility analysis of differential privacy in machine learning},
  author={Jarin, Ismat and Eshete, Birhanu},
  booktitle={Proceedings of the Twelfth ACM Conference on Data and Application Security and Privacy},
  pages={41--52},
  year={2022}
}

@inproceedings{wei2023dpmlbench,
  title={Dpmlbench: Holistic evaluation of differentially private machine learning},
  author={Wei, Chengkun and Zhao, Minghu and Zhang, Zhikun and Chen, Min and Meng, Wenlong and Liu, Bo and Fan, Yuan and Chen, Wenzhi},
  booktitle={Proceedings of the 2023 ACM SIGSAC Conference on Computer and Communications Security},
  pages={2621--2635},
  year={2023}
}

@inproceedings{BaradadJurjoWWIT21,
  title         = {Learning to See by Looking at Noise},
  author        = {Baradad Jurjo, Manel and Wulff, Jonas and Wang, Tongzhou and Isola, Phillip and Torralba, Antonio},
  booktitle     = {Advances in Neural Information Processing Systems 34},
  series        = {NeurIPS '21},
  year          = {2021},
  pages         = {2556--2569},
  publisher     = {Curran Associates, Inc.}
}

@inproceedings{IoffeS15,
  author        = {Ioffe, Sergey and Szegedy, Christian},
  title         = {Batch Normalization: Accelerating Deep Network Training by Reducing Internal Covariate Shift},
  booktitle     = {Proceedings of the 32nd International Conference on Machine Learning},
  series        = {ICML '15},
  year          = {2015},
  pages         = {448--456},
  publisher     = {JMLR, Inc.}
}

@inproceedings{JungLNRSS20,
  author        = {Jung, Christopher and Ligett, Katrina and Neel, Seth and Roth, Aaron and Sharifi-Malvajerdi, Saeed and Shenfeld, Moshe},
  title         = {A New Analysis of Differential Privacy’s Generalization Guarantees},
  booktitle     = {Proceedings of the 11th Conference on Innovations in Theoretical Computer Science},
  series        = {ITCS '20},
  year          = {2020},
  pages         = {31:1--31:17},
  publisher     = {Schloss Dagstuhl--Leibniz-Zentrum fuer Informatik},
  address       = {Dagstuhl, Germany}
}

@inproceedings{KingmaB15,
  author        = {Kingma, Diederik P and Ba, Jimmy},
  title         = {Adam: A Method for Stochastic Optimization},
  booktitle     = {Proceedings of the 3rd International Conference on Learning Representations},
  series        = {ICLR '15},
  year          = {2015}
}

@article{YousefpourSSTPMNGBZCM21,
  title         = {Opacus: User-Friendly Differential Privacy Library in {PyTorch}},
  author        = {Yousefpour, Ashkan and Shilov, Igor and Sablayrolles, Alexandre and Testuggine, Davide and Prasad, Karthik and Malek, Mani and Nguyen, John and Gosh, Sayan and Bharadwaj, Akash and Zhao, Jessica and Cormode, Graham and Mironov, Ilya},
  journal       = {arXiv preprint arXiv:2109.12298},
  year          = {2021}
}

@inproceedings{PaszkeGMLBCKLGADKYDRTCSFBC19,
  title         = {{PyTorch}: An Imperative Style, High-Performance Deep Learning Library},
  author        = {Paszke, Adam and Gross, Sam and Massa, Francisco and Lerer, Adam and Bradbury, James and Chanan, Gregory and Killeen, Trevor and Lin, Zeming and Gimelshein, Natalia and Antiga, Luca and Desmaison, Alban and K\"{o}pf, Andreas and Yang, Edward and DeVito, Zach and Raison, Martin and Tejani, Alykhan and Chilamkurthy, Sasank and Steiner, Benoit and Fang, Lu and Bai, Junjie and Chintala, Soumith},
  booktitle     = {Advances in Neural Information Processing Systems 32},
  series        = {NeurIPS '19},
  year          = {2019},
  pages         = {8026--8037},
  publisher     = {Curran Associates, Inc.}
}

@article{HarderJSP23,
  author        = {Harder, Frederik and Jalali, Milad and Sutherland, Danica J and Park, Mijung},
  title         = {Pre-trained Perceptual Features Improve Differentially Private Image Generation},
  journal       = {Transactions on Machine Learning Research},
  year          = {2023}
}

@article{DockhornCVK23,
  author        = {Dockhorn, Tim and Cao, Tianshi and Vahdat, Arash and Kreis, Karsten},
  title         = {Differentially Private Diffusion Models},
  journal       = {Transactions on Machine Learning Research},
  year          = {2023}
}

@article{BeaulieuJonesWWLBBG19,
  author        = {{Beaulieu-Jones}, Brett K and Wu, Zhiwei Steven and Williams, Chris and Lee, Ran and Bhavnani, Sanjeev P and Byrd, James Brian and Greene, Casey S},
  title         = {Privacy-Preserving Generative Deep Neural Networks Support Clinical Data Sharing},
  journal       = {Circulation: Cardiovascular Quality and Outcomes},
  volume        = {12},
  number        = {7},
  pages         = {e005122},
  year          = {2019},
  publisher     = {American Heart Association}
}

@inproceedings{CaoBVFK21,
  title         = {Don't Generate Me: Training Differentially Private Generative Models with Sinkhorn Divergence},
  author        = {Cao, Tianshi and Bie, Alex and Vahdat, Arash and Fidler, Sanja and Kreis, Karsten},
  booktitle     = {Advances in Neural Information Processing Systems 34},
  series        = {NeurIPS '21},
  year          = {2021},
  pages         = {12480--12492},
  publisher     = {Curran Associates, Inc.}
}

@article{XieLWWZ18,
  title         = {Differentially Private Generative Adversarial Network},
  author        = {Xie, Liyang and Lin, Kaixiang and Wang, Shu and Wang, Fei and Zhou, Jiayu},
  journal       = {arXiv preprint arXiv:1802.06739},
  year          = {2018}
}

@inproceedings{DBLP:conf/iclr/DosovitskiyB0WZ21,
  author       = {Alexey Dosovitskiy and
                  Lucas Beyer and
                  Alexander Kolesnikov and
                  Dirk Weissenborn and
                  Xiaohua Zhai and
                  Thomas Unterthiner and
                  Mostafa Dehghani and
                  Matthias Minderer and
                  Georg Heigold and
                  Sylvain Gelly and
                  Jakob Uszkoreit and
                  Neil Houlsby},
  title        = {An Image is Worth 16x16 Words: Transformers for Image Recognition
                  at Scale},
  booktitle    = {{ICLR}},
  publisher    = {OpenReview.net},
  year         = {2021}
}

@article{LebedaRKS24,
  title         = {Avoiding Pitfalls for Privacy Accounting of Subsampled Mechanisms under Composition},
  author        = {Lebeda, Christian Janos and Regehr, Matthew and Kamath, Gautam and Steinke, Thomas},
  journal       = {arXiv preprint arXiv:2405.20769},
  year          = {2024}
}

@inproceedings{ChuaGKKMSZ24,
  title         = {How Private are {DP-SGD} Implementations?},
  author        = {Chua, Lynn and Ghazi, Badih and Kamath, Pritish and Kumar, Ravi and Manurangsi, Pasin and Sinha, Amer and Zhang, Chiyuan},
  booktitle     = {Proceedings of the 41st International Conference on Machine Learning},
  series        = {ICML '24},
  year          = {2024},
  publisher     = {JMLR, Inc.}
}

@article{PonomarevaHKXDMVCT23,
  title         = {How to {DP}-fy {ML}: A Practical Guide to Machine Learning with Differential Privacy},
  author        = {Ponomareva, Natalia and Hazimeh, Hussein and Kurakin, Alex and Xu, Zheng and Denison, Carson and McMahan, H Brendan and Vassilvitskii, Sergei and Chien, Steve and Thakurta, Abhradeep Guha},
  journal       = {Journal of Artificial Intelligence Research},
  volume        = {77},
  pages         = {1113--1201},
  year          = {2023}
}

@inproceedings{KoskelaJH20,
  author        = {Koskela, Antti and J{\"a}lk{\"o}, Joonas and Honkela, Antti},
  title         = {Computing Tight Differential Privacy Guarantees using {FFT}},
  booktitle     = {Proceedings of the 23rd International Conference on Artificial Intelligence and Statistics},
  series        = {AISTATS '20},
  year          = {2020},
  publisher     = {JMLR, Inc.},
  pages         = {2560--2569}
}

@inproceedings{GopiLW21,
  title         = {Numerical Composition of Differential Privacy},
  author        = {Gopi, Sivakanth and Lee, Yin Tat and Wutschitz, Lukas},
  booktitle     = {Advances in Neural Information Processing Systems 34},
  series        = {NeurIPS '21},
  year          = {2021},
  pages         = {11631--11642},
  publisher     = {Curran Associates, Inc.}
}

@inproceedings{RedbergKW23,
  title         = {Improving the Privacy and Practicality of Objective Perturbation for Differentially Private Linear Learners},
  author        = {Redberg, Rachel and Koskela, Antti and Wang, Yu-Xiang},
  booktitle     = {Advances in Neural Information Processing Systems 36},
  series        = {NeurIPS '23},
  year          = {2023},
  pages         = {13819--13853},
  publisher     = {Curran Associates, Inc.}
}

@article{BieKZ23,
  author        = {Bie, Alex and Kamath, Gautam and Zhang, Guojun},
  title         = {Private {GAN}s, Revisited},
  journal       = {Transactions on Machine Learning Research},
  year          = {2023}
}

@inproceedings{BirhaneP21,
  author        = {Birhane, Abeba and Prabhu, Vinay Uday},
  title         = {Large Image Datasets: A Pyrrhic Win for Computer Vision?},
  booktitle     = {2021 IEEE Winter Conference on Applications of Computer Vision},
  series        = {WACV '21},
  year          = {2021},
  pages         = {1536--1546},
  publisher     = {IEEE}
}

@misc{LeCunCB10,
  author        = {LeCun, Yann and Cortes, Corinna and Burges, Chris},
  title         = {{MNIST} handwritten digit database},
  year          = {2010}
}

@inproceedings{TramerKC24,
  title         = {Position: Considerations for Differentially Private Learning with Large-Scale Public Pretraining},
  author        = {Tram{\`e}r, Florian and Kamath, Gautam and Carlini, Nicholas},
  booktitle     = {Proceedings of the 41st International Conference on Machine Learning},
  series        = {ICML '24},
  year          = {2024},
  publisher     = {JMLR, Inc.}
}

@inproceedings{FrederiksonJR15,
  title         = {Model Inversion Attacks that Exploit Confidence Information and Basic Countermeasures},
  author        = {Fredrikson, Matt and Jha, Somesh and Ristenpart, Thomas},
  booktitle     = {Proceedings of the 2015 ACM Conference on Computer and Communications Security},
  series        = {CCS '15},
  year          = {2015},
  pages         = {1322--1333},
  publisher     = {ACM}
}

@inproceedings{PapernotTSCE21,
  author        = {Papernot, Nicolas and Thakurta, Abhradeep and Song, Shuang and Chien, Steve and Erlingsson, {\'U}lfar},
  title         = {Tempered Sigmoid Activations for Deep Learning with Differential Privacy},
  booktitle     = {Proceedings of the Thirty-Fifth AAAI Conference on Artificial Intelligence},
  series        = {AAAI '21},
  pages         = {9312--9321},
  year          = {2021}
}

@inproceedings{SomepalliSGGG23,
  author        = {Somepalli, Gowthami and Singla, Vasu and Goldblum, Micah and Geiping, Jonas and Goldstein, Tom},
  title         = {Diffusion art or digital forgery? investigating data replication in diffusion models},
  booktitle     = {Proceedings of the 2023 IEEE Computer Society Conference on Computer Vision and Pattern Recognition},
  series        = {CVPR '23},
  year          = {2023},
  publisher     = {IEEE Computer Society},
  pages         = {6048--6058}
}

@inproceedings{CarliniHNJSTBIW23,
  author        = {Carlini, Nicolas and Hayes, Jamie and Nasr, Milad and Jagielski, Matthew and Sehwag, Vikash and Tramer, Florian and Balle, Borja and Ippolito, Daphne and Wallace, Eric},
  title         = {Extracting Training Data from Diffusion Models},
  booktitle     = {32nd USENIX Security Symposium},
  series        = {USENIX Security '23},
  year          = {2023},
  pages         = {5253--5270},
  publisher     = {USENIX Association}
}

@inproceedings{schuhmann2022laionb,
  title={{LAION}-5B: An open large-scale dataset for training next generation image-text models},
  author={Christoph Schuhmann and
          Romain Beaumont and
          Richard Vencu and
          Cade W Gordon and
          Ross Wightman and
          Mehdi Cherti and
          Theo Coombes and
          Aarush Katta and
          Clayton Mullis and
          Mitchell Wortsman and
          Patrick Schramowski and
          Srivatsa R Kundurthy and
          Katherine Crowson and
          Ludwig Schmidt and
          Robert Kaczmarczyk and
          Jenia Jitsev},
  booktitle={Thirty-sixth Conference on Neural Information Processing Systems Datasets and Benchmarks Track},
  year={2022},
  url={https://openreview.net/forum?id=M3Y74vmsMcY}
}

@inproceedings{yuan2021large,
  title={Large-scale robust deep auc maximization: A new surrogate loss and empirical studies on medical image classification},
  author={Yuan, Zhuoning and Yan, Yan and Sonka, Milan and Yang, Tianbao},
  booktitle={Proceedings of the IEEE/CVF International Conference on Computer Vision},
  pages={3040--3049},
  year={2021}
}

@article{DeBHSB22,
  title         = {Unlocking High-Accuracy Differentially Private Image Classification through Scale},
  author        = {De, Soham and Berrada, Leonard and Hayes, Jamie and Smith, Samuel L and Balle, Borja},
  journal       = {arXiv preprint arXiv:2204.13650},
  year          = {2022}
}

@article{BerradaDSHSSKSB23,
  title         = {Unlocking Accuracy and Fairness in Differentially Private Image Classification},
  author        = {Berrada, Leonard and De, Soham and Shen, Judy Hanwen and Hayes, Jamie and Stanforth, Robert and Stutz, David and Kohli, Pushmeet and Smith, Samuel L and Balle, Borja},
  journal       = {arXiv preprint arXiv:2308.10888},
  year          = {2023}
}

@misc{eyepacs,
  author = {Eyepacs},
  title = {Diabetic Retinopathy Detection},
  year = {2015},
  note = {data retrieved from Kaggle, 
          \url{https://www.kaggle.com/c/diabetic-retinopathy-detection}},
}

@inproceedings{radford2021learning,
  title={Learning transferable visual models from natural language supervision},
  author={Radford, Alec and Kim, Jong Wook and Hallacy, Chris and Ramesh, Aditya and Goh, Gabriel and Agarwal, Sandhini and Sastry, Girish and Askell, Amanda and Mishkin, Pamela and Clark, Jack and others},
  booktitle={International conference on machine learning},
  pages={8748--8763},
  year={2021},
  organization={PMLR}
}

@inproceedings{AbadiCGMMTZ16,
  title         = {Deep Learning with Differential Privacy},
  author        = {Abadi, Martin and Chu, Andy and Goodfellow, Ian and McMahan, H Brendan and Mironov, Ilya and Talwar, Kunal and Zhang, Li},
  booktitle     = {Proceedings of the 2016 ACM Conference on Computer and Communications Security},
  series        = {CCS '16},
  year          = {2016},
  publisher     = {ACM},
  pages         = {308--318},
  address       = {New York, NY, USA}
}

@article{AnilGGKM21,
  title         = {Large-Scale Differentially Private {BERT}},
  author        = {Anil, Rohan and Ghazi, Badih and Gupta, Vineet and Kumar, Ravi and Manurangsi, Pasin},
  journal       = {arXiv preprint arXiv:2108.01624},
  year          = {2021}
}

@article{bassily2020stability,
  title={Stability of stochastic gradient descent on nonsmooth convex losses},
  author={Bassily, Raef and Feldman, Vitaly and Guzm{\'a}n, Crist{\'o}bal and Talwar, Kunal},
  journal={Advances in Neural Information Processing Systems},
  volume={33},
  pages={4381--4391},
  year={2020}
}

@inproceedings{IrvinRKYCCMHBSSMHSJLLPLN19,
  title         = {Chexpert: A large chest radiograph dataset with uncertainty labels and expert comparison},
  author        = {Irvin, Jeremy and Rajpurkar, Pranav and Ko, Michael and Yu, Yifan and Ciurea-Ilcus, Silviana and Chute, Chris and Marklund, Henrik and Haghgoo, Behzad and Ball, Robyn and Shpanskaya, Katie and Seekins, Jayne and Mong, David A. and Halabi, Safwan S. and Sandberg, Jesse K. and Jones, Ricky and Larson, David B. and Langlotz, Curtis P. and Patel, Bhavik N. and Lungren, Matthew P. and Ng, Andrew Y.},
  booktitle     = {Proceedings of the Thirty-Third AAAI Conference on Artificial Intelligence},
  series        = {AAAI '19},
  year          = {2019},
  pages         = {590--597}
}

@inproceedings{BassilyST14,
  author        = {Bassily, Raef and Smith, Adam and Thakurta, Abhradeep},
  title         = {Private Empirical Risk Minimization: Efficient Algorithms and Tight Error Bounds},
  booktitle     = {Proceedings of the 55th Annual IEEE Symposium on Foundations of Computer Science},
  series        = {FOCS '14},
  year          = {2014},
  pages         = {464--473},
  publisher     = {IEEE Computer Society},
  address       = {Washington, DC, USA},
}

@inproceedings{CarliniTWJHLRBSEOR21,
  author        = {Carlini, Nicholas and Tram\`er, Florian and Wallace, Eric and Jagielski, Matthew and Herbert-Voss, Ariel and Lee, Katherine and Roberts, Adam and Brown, Tom and Song, Dawn and Erlingsson, Ulfar and Oprea, Alina and Raffel, Colin},
  title         = {Extracting Training Data from Large Language Models},
  booktitle     = {30th USENIX Security Symposium},
  series        = {USENIX Security '21},
  year          = {2021},
  pages         = {2633--2650},
  publisher     = {USENIX Association}
}

@article{ChaudhuriMS11,
  author        = {Chaudhuri, Kamalika and Monteleoni, Claire and Sarwate, Anand D.},
  title         = {Differentially Private Empirical Risk Minimization},
  journal       = {Journal of Machine Learning Research},
  volume        = {12},
  number        = {29},
  pages         = {1069--1109},
  year          = {2011},
  publisher     = {JMLR, Inc.}
}

@inproceedings{DengDSLLF09,
  author        = {Deng, Jia and Dong, Wei and Socher, Richard and Li, Li-Jia and Li, Kai and Fei-Fei, Li},
  title         = {Imagenet: A Large-Scale Hierarchical Image Database},
  booktitle     = {Proceedings of the 2009 IEEE Computer Society Conference on Computer Vision and Pattern Recognition},
  series        = {CVPR '09},
  year          = {2009},
  pages         = {248--255},
  publisher     = {IEEE Computer Society},
  address       = {Washington, DC, USA}
}

@inproceedings{DworkKMMN06,
  author        = {Dwork, Cynthia and Kenthapadi, Krishnaram and McSherry, Frank and Mironov, Ilya and Naor, Moni},
  title         = {Our Data, Ourselves: Privacy via Distributed Noise Generation},
  booktitle     = {Proceedings of the 24th Annual International Conference on the Theory and Applications of Cryptographic Techniques},
  series        = {EUROCRYPT '06},
  year          = {2006},
  pages         = {486--503},
  publisher     = {Springer},
  address       = {Berlin, Heidelberg}
}

@inproceedings{DworkMNS06,
  author        = {Dwork, Cynthia and McSherry, Frank and Nissim, Kobbi and Smith, Adam},
  title         = {Calibrating Noise to Sensitivity in Private Data Analysis},
  booktitle     = {Proceedings of the 3rd Conference on Theory of Cryptography},
  series        = {TCC '06},
  year          = {2006},
  pages         = {265--284},
  publisher     = {Springer},
  address       = {Berlin, Heidelberg}
}

@article{HomerSRDTMPSNC08,
  author        = {Homer, Nils and Szelinger, Szabolcs and Redman, Margot and Duggan, David and Tembe, Waibhav and Muehling, Jill and Pearson, John V. and Stephan, Dietrich A. and Nelson, Stanley F. and Craig, David W.},
  title         = {Resolving Individuals Contributing Trace Amounts of {DNA} to Highly Complex Mixtures using High-Density {SNP} Genotyping Microarrays},
  journal       = {PLoS Genetics},
  volume        = {4},
  number        = {8},
  pages         = {1--9},
  year          = {2008},
  publisher     = {Public Library of Science}
}

@inproceedings{IyengarNSTTW19,
  author        = {Iyengar, Roger and Near, Joseph P and Song, Dawn and Thakkar, Om and Thakurta, Abhradeep and Wang, Lun},
  title         = {Towards Practical Differentially Private Convex Optimization},
  booktitle     = {Proceedings of the 40th IEEE Symposium on Security and Privacy},
  series        = {SP '19},
  year          = {2019},
  pages         = {299--316},
  publisher     = {IEEE Computer Society},
  address       = {Washington, DC, USA}
}

@inproceedings{KiferST12,
  author        = {Kifer, Daniel and Smith, Adam and Thakurta, Abhradeep},
  title         = {Private Convex Empirical Risk Minimization and High-Dimensional Regression},
  booktitle     = {Proceedings of the 25th Annual Conference on Learning Theory},
  series        = {COLT '12},
  year          = {2012},
  pages         = {25.1--25.40}
}

@misc{Krizhevsky09,
  author        = {Krizhevsky, Alex},
  title         = {Learning Multiple Layers of Features from Tiny Images},
  year          = {2009}
}

@inproceedings{Mironov17,
  author        = {Mironov, Ilya},
  title         = {R{\'e}nyi Differential Privacy},
  booktitle     = {Proceedings of the 30th IEEE Computer Security Foundations Symposium},
  series        = {CSF '17},
  year          = {2017},
  pages         = {263--275},
  publisher     = {IEEE Computer Society},
  address       = {Washington, DC, USA},
}

@article{MironovTZ19,
  title         = {R{\'e}nyi Differential Privacy of the Sampled Gaussian Mechanism},
  author        = {Mironov, Ilya and Talwar, Kunal and Zhang, Li},
  journal       = {arXiv preprint arXiv:1908.10530},
  year          = {2019}
}

@inproceedings{ShokriSSS17,
  author        = {Shokri, Reza and Stronati, Marco and Song, Congzheng and Shmatikov, Vitaly},
  title         = {Membership Inference Attacks Against Machine Learning Models},
  booktitle     = {Proceedings of the 38th IEEE Symposium on Security and Privacy},
  series        = {SP '17},
  year          = {2017},
  pages         = {3--18},
  publisher     = {IEEE Computer Society},
  address       = {Washington, DC, USA},
}

@inproceedings{SongCS13,
  author        = {Song, Shuang and Chaudhuri, Kamalika and Sarwate, Anand D},
  title         = {Stochastic Gradient Descent with Differentially Private Updates},
  booktitle     = {Proceedings of the 2013 IEEE Global Conference on Signal and Information Processing},
  series        = {GlobalSIP '13},
  year          = {2013},
  pages         = {245--248},
  publisher     = {IEEE Computer Society},
  address       = {Washington, DC, USA}
}

@article{TalwarTZ14,
  title         = {Private Empirical Risk Minimization Beyond the Worst Case: The Effect of the Constraint Set Geometry},
  author        = {Talwar, Kunal and Thakurta, Abhradeep and Zhang, Li},
  journal       = {arXiv preprint arXiv:1411.5417},
  year          = {2014}
}

@inproceedings{TramerB21,
  author        = {Tram{\`e}r, Florian and Boneh, Dan},
  title         = {Differentially Private Learning Needs Better Features (or Much More Data)},
  booktitle     = {Proceedings of the 9th International Conference on Learning Representations},
  series        = {ICLR '21},
  year          = {2021}
}

@inproceedings{LiTLH22,
  author        = {Li, Xuechen and Tram\`er, Florian and Liang, Percy and Hashimoto, Tatsunori},
  title         = {Large Language Models Can Be Strong Differentially Private Learners},
  booktitle     = {Proceedings of the 10th International Conference on Learning Representations},
  series        = {ICLR '22},
  year          = {2022}
}

@inproceedings{YuNBGIKKLMWYZ22,
  author        = {Yu, Da and Naik, Saurabh and Backurs, Arturs and Gopi, Sivakanth and Inan, Huseyin A and Kamath, Gautam and Kulkarni, Janardhan and Lee, Yin Tat and Manoel, Andre and Wutschitz, Lukas and Yekhanin, Sergey and Zhang, Huishuai},
  title         = {Differentially Private Fine-tuning of Language Models},
  booktitle     = {Proceedings of the 10th International Conference on Learning Representations},
  series        = {ICLR '22},
  year          = {2022}
}

@article{Oyallon14,
  author       = {Edouard Oyallon and
                  St{\'{e}}phane Mallat},
  title        = {Deep Roto-Translation Scattering for Object Classification},
  journal      = {CoRR},
  volume       = {abs/1412.8659},
  year         = {2014},
  url          = {http://arxiv.org/abs/1412.8659},
  eprinttype    = {arXiv},
  bibsource    = {dblp computer science bibliography, https://dblp.org}
}

@inproceedings{yuan2023libauc,
	title={LibAUC: A Deep Learning Library for X-risk Optimization},
	author={Zhuoning Yuan and Dixian Zhu and Zi-Hao Qiu and Gang Li and Xuanhui Wang and Tianbao Yang},
	booktitle={29th SIGKDD Conference on Knowledge Discovery and Data Mining},
	year={2023}
	}

@article{Wu18,
  author       = {Yuxin Wu and
                  Kaiming He},
  title        = {Group Normalization},
  journal      = {CoRR},
  volume       = {abs/1803.08494},
  year         = {2018},
  url          = {http://arxiv.org/abs/1803.08494},
  eprinttype    = {arXiv},
  eprint       = {1803.08494},
  bibsource    = {dblp computer science bibliography, https://dblp.org}
}

@misc{raghu2019transfusion,
      title={Transfusion: Understanding Transfer Learning for Medical Imaging}, 
      author={Maithra Raghu and Chiyuan Zhang and Jon Kleinberg and Samy Bengio},
      year={2019},
      eprint={1902.07208},
      archivePrefix={arXiv},
      primaryClass={cs.CV}
}

@article{brock2021,
  author       = {Andrew Brock and
                  Soham De and
                  Samuel L. Smith},
  title        = {Characterizing signal propagation to close the performance gap in
                  unnormalized ResNets},
  journal      = {CoRR},
  volume       = {abs/2101.08692},
  year         = {2021},
  url          = {https://arxiv.org/abs/2101.08692},
  eprinttype    = {arXiv},
  eprint       = {2101.08692},
  bibsource    = {dblp computer science bibliography, https://dblp.org}
}

@article{Voets_2019,
   title={Reproduction study using public data of: Development and validation of a deep learning algorithm for detection of diabetic retinopathy in retinal fundus photographs},
   volume={14},
   ISSN={1932-6203},
   url={http://dx.doi.org/10.1371/journal.pone.0217541},
   DOI={10.1371/journal.pone.0217541},
   number={6},
   journal={PLOS ONE},
   publisher={Public Library of Science (PLoS)},
   author={Voets, Mike and Møllersen, Kajsa and Bongo, Lars Ailo},
   editor={Ortega-Martorell, Sandra},
   year={2019},
   month=jun, pages={e0217541} }

@misc{dörmann2021noise,
      title={Not all noise is accounted equally: How differentially private learning benefits from large sampling rates}, 
      author={Friedrich Dörmann and Osvald Frisk and Lars Nørvang Andersen and Christian Fischer Pedersen},
      year={2021},
      eprint={2110.06255},
      archivePrefix={arXiv},
      primaryClass={cs.LG}
}

@misc{he2015deep,
      title={Deep Residual Learning for Image Recognition}, 
      author={Kaiming He and Xiangyu Zhang and Shaoqing Ren and Jian Sun},
      year={2015},
      eprint={1512.03385},
      archivePrefix={arXiv},
      primaryClass={cs.CV}
}

@article{DBLP:journals/corr/ZagoruykoK16,
  author       = {Sergey Zagoruyko and
                  Nikos Komodakis},
  title        = {Wide Residual Networks},
  journal      = {CoRR},
  volume       = {abs/1605.07146},
  year         = {2016},
  url          = {http://arxiv.org/abs/1605.07146},
  eprinttype    = {arXiv},
  eprint       = {1605.07146},
  bibsource    = {dblp computer science bibliography, https://dblp.org}
}

@misc{brock2021highperformancelargescaleimagerecognition,
      title={High-Performance Large-Scale Image Recognition Without Normalization}, 
      author={Andrew Brock and Soham De and Samuel L. Smith and Karen Simonyan},
      year={2021},
      eprint={2102.06171},
      archivePrefix={arXiv},
      primaryClass={cs.CV},
      url={https://arxiv.org/abs/2102.06171}, 
}

@article{DBLP:journals/corr/ChrabaszczLH17,
  author       = {Patryk Chrabaszcz and
                  Ilya Loshchilov and
                  Frank Hutter},
  title        = {A Downsampled Variant of ImageNet as an Alternative to the {CIFAR}
                  datasets},
  journal      = {CoRR},
  volume       = {abs/1707.08819},
  year         = {2017},
  url          = {http://arxiv.org/abs/1707.08819},
  eprinttype    = {arXiv},
  eprint       = {1707.08819},
  bibsource    = {dblp computer science bibliography, https://dblp.org}
}

@article{PolyakEMA,
author = {Polyak, Boris and Juditsky, Anatoli},
year = {1992},
month = {07},
pages = {838-855},
title = {Acceleration of Stochastic Approximation by Averaging},
volume = {30},
journal = {SIAM Journal on Control and Optimization},
doi = {10.1137/0330046}
}

@misc{tao2022benchmarkingdifferentiallyprivatesynthetic,
      title={Benchmarking Differentially Private Synthetic Data Generation Algorithms}, 
      author={Yuchao Tao and Ryan McKenna and Michael Hay and Ashwin Machanavajjhala and Gerome Miklau},
      year={2022},
      eprint={2112.09238},
      archivePrefix={arXiv},
      primaryClass={cs.CR},
      url={https://arxiv.org/abs/2112.09238}, 
}

@misc{tang2023differentiallyprivateimageclassification,
      title={Differentially Private Image Classification by Learning Priors from Random Processes}, 
      author={Xinyu Tang and Ashwinee Panda and Vikash Sehwag and Prateek Mittal},
      year={2023},
      eprint={2306.06076},
      archivePrefix={arXiv},
      primaryClass={cs.CV},
      url={https://arxiv.org/abs/2306.06076}, 
}
